\documentclass[a4paper, 11pt]{article}

\usepackage[utf8]{inputenc}
\usepackage[english]{babel}
\usepackage{graphicx}
\usepackage{amsfonts,amssymb}
\usepackage{url}
\usepackage{hyperref}
\usepackage{booktabs}

\usepackage{natbib}
\usepackage{comment}
\usepackage{amsmath}

\newcommand\numberthis{\addtocounter{equation}{1}\tag{\theequation}}

\newcommand{\bbRm}[1]{\mathrm{\mathbb{#1}}}

\def\vd{\mathbf{d}}
\def\ve{\mathbf{e}}
\def\vo{\mathbf{o}}
\def\vx{\mathbf{x}}
\def\vy{\mathbf{y}}
\def\vz{\mathbf{z}}
\def\mI{\mathbf{I}}
\def\mU{\mathbf{U}}
\def\grad{\nabla}
\def\hess{\nabla^2}
\def\R{\mathbb{R}}

\def\NV{{\mathcal N}}

\def\mX{{\mathbf{X}}}

\def\mW{{\mathbf{W}}}

\def\cF{{\mathcal F}}

\def\cJ{{\mathcal J}}

\def\Diag{{\mathop{\bbRm Diag}}}

\def\nHidOne{\textsc{1Hid}}
\def\nSymOne{\textsc{1Sym}}
\def\nSymThree{\textsc{3Sym}}
\def\nSymFive{\textsc{5Sym}}
\def\nSymSeven{\textsc{7Sym}}
\newtheorem{rem}{Remark}

\begin{document}

\title{An Additive Autoencoder for Dimension Estimation\footnote{Manuscript submitted to review}}

\author{Kärkkäinen, Tommi and Hänninen, Jan \\
tommi.karkkainen@jyu.fi, jan.p.hanninen@jyu.fi \\
Faculty of Information Technology, University of Jyväskylä, Finland}

\date{\today}

\maketitle

\begin{abstract}
An additive autoencoder for dimension reduction, which is composed of a serially performed bias estimation, linear trend estimation, and nonlinear residual estimation, is proposed and analyzed. Computational experiments confirm that an autoencoder of this form, with only a shallow network to encapsulate the nonlinear behavior, is able to identify an intrinsic dimension of a dataset with a low autoencoding error. This observation leads to an investigation in which shallow and deep network structures, and how they are trained, are compared. We conclude that the deeper network structures obtain lower autoencoding errors during the identification of the intrinsic dimension. However, the detected dimension does not change compared to a shallow network.
\end{abstract}


\section{Introduction}
\label{sec:intro}

Dimension reduction is one of the typical data transformation techniques used in data mining. Both linear and nonlinear techniques can be used to transform a set of observations into a smaller dimension \citep{burges2010dimension}. A specific and highly popular set of nonlinear methods are provided with autoencoders, AE \citep{schmidhuber2015deep}, which by using the original inputs as targets integrate unsupervised and supervised learning for dimension reduction. The main purpose of this paper is to propose and thoroughly test an autoencoding model, which comprises an additive combination of linear and nonlinear dimension reduction techniques through serially performed bias estimation, linear trend estimation, and nonlinear residual estimation. Preliminary, limited investigations of such a model structure have been reported in \citep{KarRas2020,KarPN2022}.

With the proposed autoencoding model, we consider its ability to estimate the intrinsic dimensionality of data \citep{fukunaga1971algorithm,camastra2003data, lee2007nonlinear}. According to \cite{fukunaga1982}, the intrinsic dimension can be defined as the size of the lower-dimension manifold where data lies without information loss. With linear principal component analysis (PCA), this loss can be measured with the explained variance which is measured by the eigenvalues of the covariance matrix \cite{Jolliffe2002}. Indeed, the use of an autoencoder to estimate the intrinsic dimension can be considered a nonlinear extension of the projection method based on PCA \citep{facco2017estimating}. However, in the nonlinear case measures for characterizing the essential information and how this is used to reduce the dimensionality vary \citep{camastra2016intrinsic,navarro2017empirical}. 

\cite{wang2016auto} concluded that a shallow autoencoder shows the best performance when the size of the squeezing dimension is approximately around the intrinsic dimension. In this direction, techniques that are closely related to our work were proposed by \cite{bahadur2020dimension}, where the intrinsic dimension was estimated using an autoencoder with sparsity-favoring $l_1$ penalty and singular value proxies of the squeezing layer's encoding. In the experiments, the superiority of the autoencoder compared to PCA was concluded. This and the preliminary work by \cite{bahadur2019dimension} applied \emph{a priori} fixed architectures of the autoencoder and different autoencoding error measure compared to our work. Here, multiple feedforward models are used and compared, with a simple thresholding technique to detect the intrinsic dimension based on the data reconstruction error.

Interestingly, our experiments reveal that the intrinsic dimension can be identified by using only a shallow feedforward network as the nonlinear residual operator in the additive autoencoding model. This results from including the linear operator in the overall transformation and considering the unexplained residual in the original data dimension. This does not happen with the classical autoencoder (without linear term) or if an autoencoder would be used in the reduced dimension after the linear transformation. These phenomena, with the models and techniques fully specified in the subsequent sections, is illustrated in Fig. \ref{fig:IntroIllustration}. 
Therefore, in addition to exploring the capabilities of revealing the intrinsic dimension we assess the advantages of applying deeper networks as nonlinear operators. Apparently, our results diversify views on the general superiority of deeper network architectures. They are also linked to the existing challenges that researchers need to address with deep learning techniques \citep{chen2018shallowing,lathuiliere2019comprehensive,ghods2021survey}.

\def\figurescale{4.18truecm}
\def\FigsCut{-0.4truecm}
\def\CaptCut{-0.0truecm}

\begin{figure*}[t!]
\centering
\hspace*{\FigsCut}
\includegraphics[width=\figurescale]{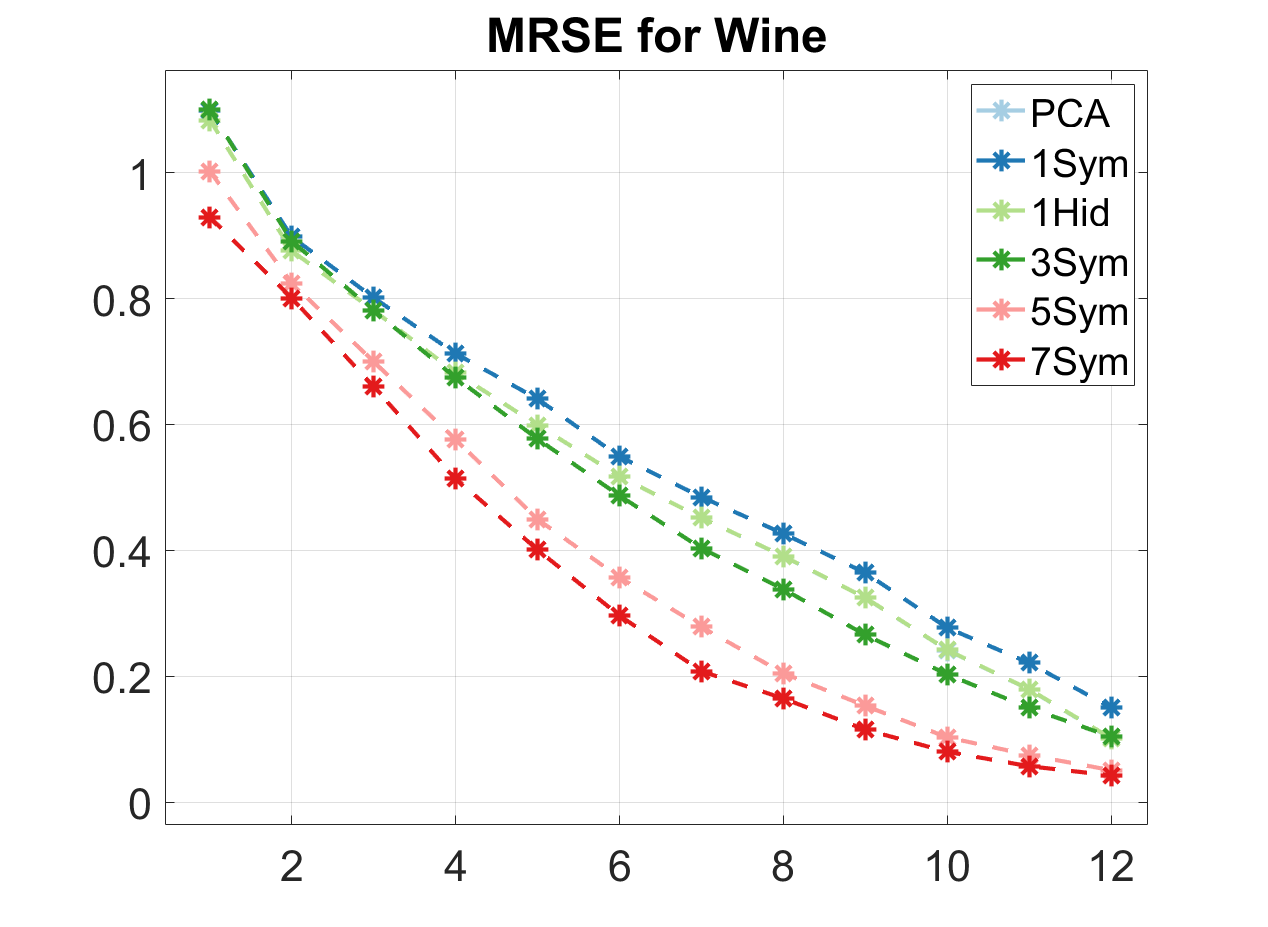}
\hspace*{\FigsCut}
\includegraphics[width=\figurescale]{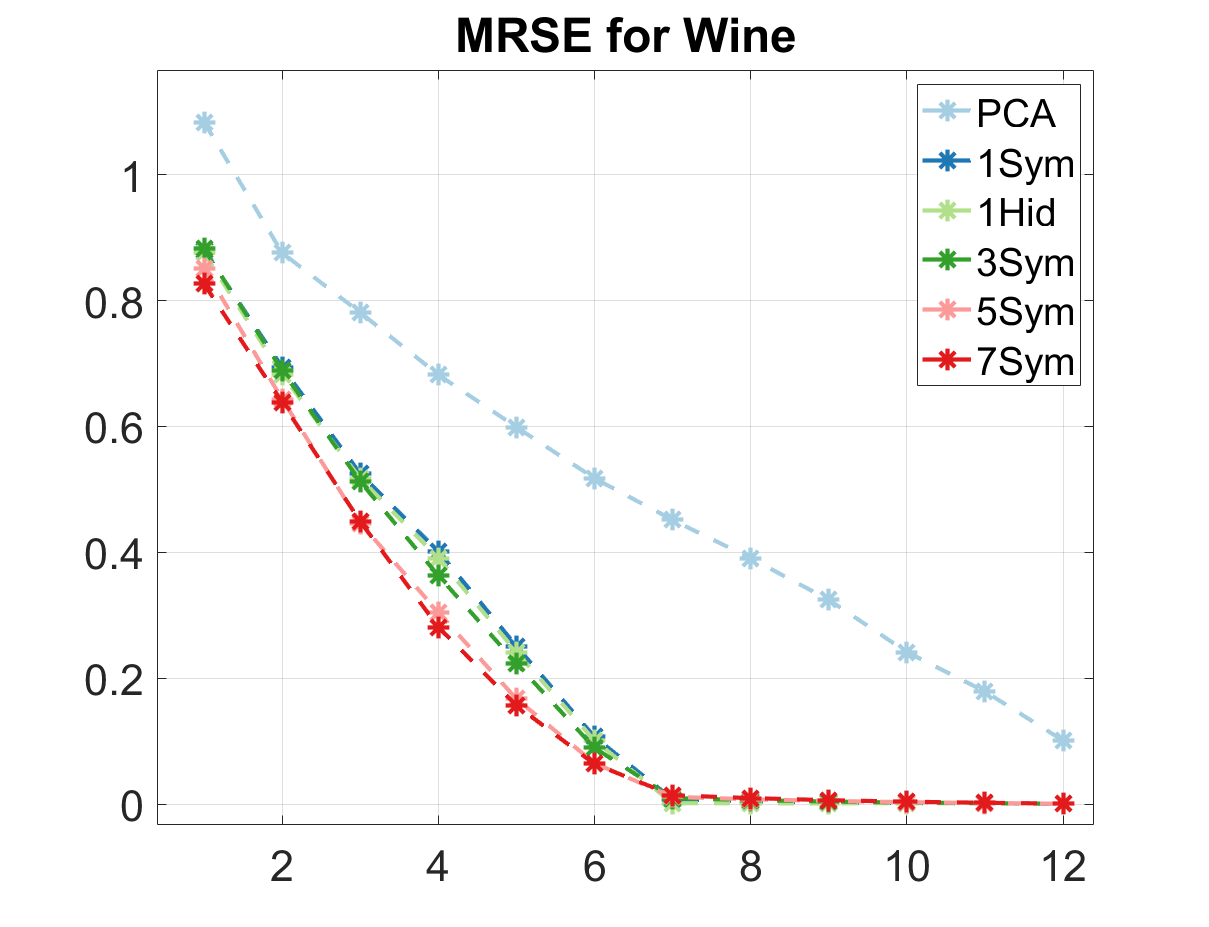}
\hspace*{\FigsCut}
\includegraphics[width=\figurescale]{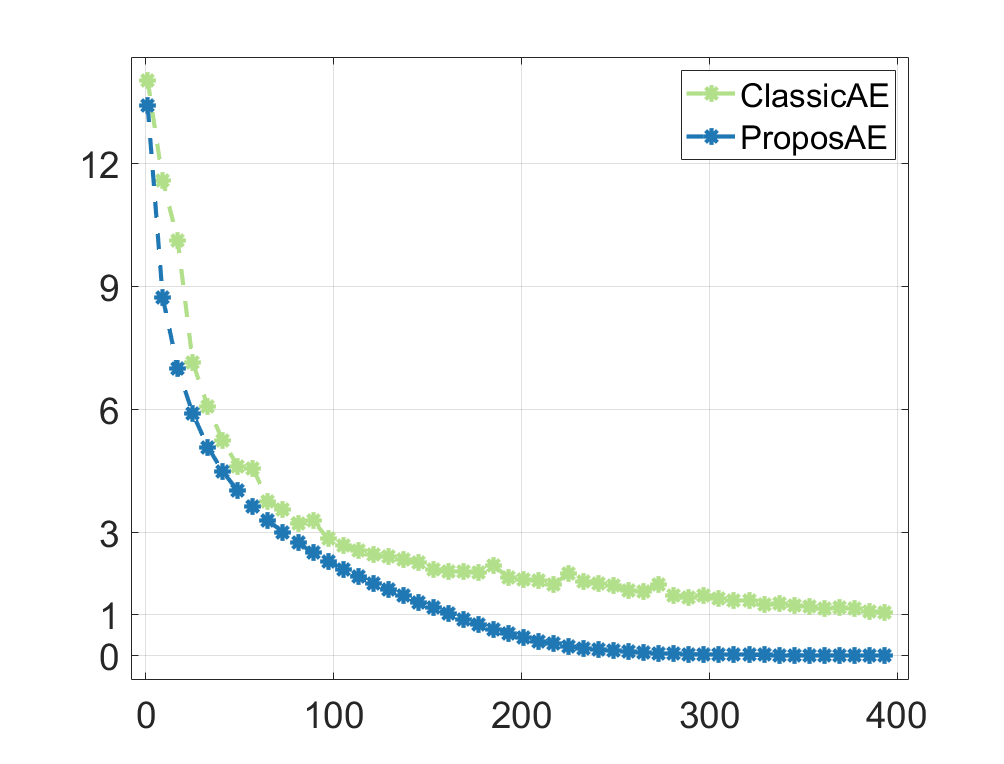}
\vspace*{\CaptCut}
\caption{Residual errors with Wine dataset for the usual autoencoders with different number of hidden layers (left), for the proposed, additive autoencoders (middle), and between these two models  with five-hidden-layers for the MNIST dataset (right). Here $x$-axis contains the squeezing dimension and $y$-axis the autoencoding (reconstruction) error. Deeper models provide lower autoencoding errors on the left, but, only with the additional linear operator as proposed here, the intrinsic dimension is revealed on the middle and right figures. Improvement due to the depth of the model is significant on the left but only moderate on the middle, where all models are stricly better than the linear PCA alone. On the right, better capability of the proposed autoencoder to encapsulate the variability of MNIST compared to the classical approach is clearly visible.
}
\label{fig:IntroIllustration}
\end{figure*}

\subsection{On Autoencoders}
\label{subsec:priorworkAutoE}

Feedforward autoencoders have a versatile history, beginning from \citep{cottrell1985learning,bourlard1988auto}. Their development as part of the evolution from shallow network models into deep learning techniques with various network architectures has been depicted in numerous large and comprehensive reviews \citep{schmidhuber2015deep,lecun2015deep,jordan2015machine}. Therefore, we only provide a brief summary of these techniques below.

Deep feedforward autoencoding was highly influenced by the seminal work of 
\cite{hinton2006reducing}. The work emphasized the importance of pretraining and the usability of stacking (i.e., layer-by-layer construction of the deeper architecture). Such techniques, by directing the determination of weights to a potential region of the search space, particularly alleviate the vanishing gradient problem (see \cite[Section 5.9]{schmidhuber2015deep} and references therein).

As summarized\textemdash for example, by \citep{liu2017survey}, many architectures and training variants for deep autoencoders (AE) have been proposed over the years:
\begin{itemize}
\item \emph{Denoising AEs (DAE)} in which noise (see \citep{ho2010objective}) is added into the training data \citep{vincent2010stacked,chen2015marginalizing,ismail2019deep,probst2020harmless,ma2020midia}.
\item \emph{Sparse AEs (SAE)} in which the number of active, non-,zero weights is minimized \citep{filzmoser2012review} (see also \cite[Section 5.6.4]{schmidhuber2015deep}).
\item \emph{Contractive AEs (CAE)} in which the reconstruction phase is penalized \citep{diallo2021deep}.
\item \emph{Separable AEs (SAE)} in which two separate deep autoencoders are applied to model signal and noise spectra \citep{sun2015unseen} (cf. Siamese neural networks that do the opposite and use shared weights \citep{ahrabian2019usage}).
\item \emph{Graph AEs (GAE)} in which graphs, or their nodes, are encoded into latent representations and back \citep{wu2021comprehensive,hou2022graphmae,yoo2022accurate}.
\item \emph{Variational AEs (VAE)} which are deep generative models that utilize Bayesian networks in learning probability distribution of data for encoding and decoding \citep{dai2018connections,burkhardt2019decoupling,zhao2021conditional,takahashi2022learning}.
\item \emph{Regularized AEs (RAE)} in which the suppression of the derivatives of the encoder and regularization function orthogonal to the manifold provide local characterization of data-generation density \citep{alain2014regularized}.
\item \emph{Multi-modal AEs} can handle and unify the processing of different data modalities \citep{janakarajan2022fully}.
\item \emph{Other AEs} typically integrate concepts and techniques from relevant areas, for instance, autoencoder bottlenecks (AE-BN) that are based on Deep Belief Networks \citep{sainath2012auto} and rough autoencoders (RAE) where rough set based neurons are used in the layers \citep{khodayar2017rough}.
\end{itemize}

A wide variety of tasks and domains has been and can be addressed with autoencoders. AEs are typically used in numerous application domains in scenarios where transfer learning can be utilized\textemdash for example, in speech processing \citep{deng2017recognizing}, time series prediction \citep{sun2018deep}, fault diagnostics \citep{sun2019sparse}, and machine vision \citep{kim2020efficient}. In addition, interesting unsupervised hybrids are provided\textemdash for example, by clustering techniques that incorporate AEs for feature transformation \citep{min2018survey,mcconville2021n2d} and unsupervised AE-based hashing methods that can be used for large-scale information retrieval \citep{zhang2021autoencoder}. Further, AEs have been used for the estimation of data distribution \citep{khajenezhad2020masked}. Use of a variational AE for joint estimation of a normal latent data distribution and the corresponding contributing dimensions has been addressed in \citep{ikeda2018estimation}. Yet another use case of autoencoders is outlier detection, which might need statistically robust first-order fitting techniques instead of second-order least-squares  \citep{gao2020rvae,karkkainen2004robust}. Data imputation has been realized using a shallow autoencoder in \citep{narayanan2002set} and, more recently, using deep autoencoders mainly for spatio-temporal data in \citep{tran2017missing,abiri2019establishing,zhao2020traffic,li2020spatiotemporal,sangeetha2020deep,ryu2020denoising}.

\subsection{Contributions and contents}
\label{subsec:contrib}

The main contribution of the paper is the derivation and evaluation of the additive autoencoding model. We provide an experimental confirmation of the new autoencoder's ability to reveal the intrinsic dimension and study the effect of model depth. Based on the similar residual idea than with the model, we also depict a simple layerwise pretraining technique. With minor role, mostly covered in the Supplementary Information (SI), we also discuss and provide an experimental illustration of the difficulties of currently popular deep learning techniques in realizing the potential of deep network models. Overall, our results suggest that current and upcoming applications in deep learning could be improved by using an explicit separation of the linear and nonlinear aspects of the data-driven model. Moreover, it might be helpful to apply more accurate training techniques.

The remainder of the paper is organized in the following manner: In Section \ref{sec:Methods}, we discuss the formalization of the proposed method as a whole.
In Section \ref{sec:Exprm}, we describe the computational experiments and summarize the main findings. In Section \ref{sec:Conclu}, we provide the overall conclusions and discussion. In the SI, we provide more background and preliminary material and, especially, report the computational experiments as a whole. Main findings solely covered on SI confirm the quality of the implementation of the methods and especially indicate that different autoencoder models, as depicted in the previous section, could be used for the nonlinear residual estimation.

\section{Methods}\label{sec:Methods}

In this section, we describe the methods used here as part of the autoencoding approach. 
In the following account, we assume that a training set of $N$ observations $\mX = \{\vx_i\}_{i=1}^N$, where $\vx_i \in \R^n$, is given.

\subsection{The autoencoding model}\label{subsec:AEModel}

In mathematical modelling, linear and nonlinear models are typically treated separately 
\citep{bellomo1994modelling}. Following \cite{KarPN2022}, according to Taylor's formula, in the neighborhood of a point $\vx_0 \in \R^n$, the value of a sufficiently smooth real-valued function $f (\vx)$ can be approximated as
\begin{align*}
f(\vx) = f(\vx_0) + \grad f(\vx_0)^T (\vx - \vx_0) 
+ \frac12 (\vx - \vx_0)^T \hess f(\vx_0) (\vx - \vx_0) + \ldots ,
\end{align*}
where $\grad f(\vx_0)$ denotes the gradient vector and $\hess f(\vx_0)$ the Hessian matrix at $\vx_0$. According to \cite[Lemma 4.1.5]{dennis1996numerical}, 
there exists $\vz \in l (\vx, \vx_0)$ (a line segment connecting the two points), such that
\begin{align*}
f(\vx) = f(\vx_0) + \grad f(\vx_0)^T (\vx - \vx_0) 
+ \frac12 (\vx - \vx_0)^T \hess f(\vz) (\vx - \vx_0).\numberthis \label{TFrom}
\end{align*}
This formula yields the sufficient condition of $\vx \in \R^n$ to be the local minimizer of a convex $f$ (whose Hessian is positive semidefinite) 
\cite[Theorem 2.2]{nocedal2006numerical}: $\grad f (x) = 0$. 
However, another interpretation of the formula above is that we can locally approximate the value of a smooth function as a sum of its bias (i.e., constant level), a linear term, and a nonlinear higher-order residual operator. This observation is the starting point for proposing an autoencoder that has exactly such an additive structure.

The bias estimation simply involves the elimination of its effect through normalization by subtracting the data mean and scaling each feature separately into the same range $[-1, 1]$ with the scaling factor $\frac2{\max(x) - \min(x)}$. Thus, we combine the mean component from z-scoring and the scaling component from min-max scaling. The reason for this is that the unit value of the standard deviation in z-scoring does not guarantee equal ranges, and min-max scaling into $[-1,1]$ does not preserve the zero mean.

In the second phase, we estimate the linear behavior of the normalized data by using the classical principal component transformation \citep{bishop1995neural}. For a zero-mean vector $\vx \in \R^n$, the transformation to a smaller-dimensional space $m < n$ spanned by $m$ principal components (PCs) is given by $\vy = \mU^T \vx$, where $\mU \in \R^{n\times m}$ consists of the $m$ most significant (as measured by the sizes of the corresponding eigenvalues) eigenvectors of the covariance matrix. Thus, because of the orthonormality of $\mU$, the unexplained residual variance of the PC coordinates (i.e., the linear trend in $\R^m$) \emph{in the original space} (see SI, Section 7) can be estimated in the following manner:
\begin{equation}\label{PCRes}
\tilde{\vx} = \vx - \mU \vy = \vx - \mU \mU^T \vx = (\mI - \mU \mU^T) \vx .
\end{equation}
This transformation is referred to as PCA. With erroneous data or data with missing values, mean and classical PCA can be replaced with their statistically robust counterparts \citep{karkkainen2015robust}.

In the third, nonlinear phase, we apply the classical fully connected feedforward autoencoder to the residual vectors in \eqref{PCRes}. As anticipated by the scaling, the $\tanh$ activation function $f(x) = \frac2{1 + \exp(-2 x)} - 1$ is used. This ensures the smoothness of the entire transformation and the optimization problem of determining the weights. The currently popular rectified linear units are nondifferentiable \citep{karkkainen2004robust} and, therefore, are not 
theoretically 
compatible with the gradient-based methods \cite[Section 6.3.1]{goodfellow2016deep}. The importance of differentiability was also noted in \citep{ghods2021survey}.

The formalism introduced by \cite{karkkainen2002mlp} is used for the compact derivation of the optimality conditions. 
By representing the layerwise activation using {\em diagonal function-matrix} $\cF = \cF (\cdot) =
\Diag \{ f_i (\cdot ) \}_{i = 1}^m$, where $f_i \equiv f$, 
the output of a feedforward network with $L$
layers and linear activation on the final layer
reads as
\begin{equation}\label{FFTrans}
\vo = \vo^L = \NV (\tilde{\vx}) = \mW^L \vo^{(L-1)},
\end{equation}
where $\vo^0 = \tilde{\vx}$ for an input vector $\tilde{\vx} \in \R^{n_0}$, and
$\vo^l = \cF (\mW^l \vo^{(l-1)}) \text{ for } l = 1, \ldots, L - 1$. 
To allow the formal adjoint transformation to be used as the decoder, we assume that $L$ is even and that the bias nodes are not included in \eqref{FFTrans}. The dimensions of the weight matrices are then given by $\dim (\mW^l) = n_l\times n_{l-1},\ l = 1, \dots, L$. In the autoencoding context, $n_L = n_0$ and $n_l, 0 < l < L,$ define the sizes (the number of neurons) of the hidden layers with the squeezing dimension $n_{L/2} < n_0$.

To determine the weights in \eqref{FFTrans}, we minimize the regularized mean least-squares cost function of the form
\begin{align*}
\cJ (\{\mW^l\}_{l=1}^L) = &\frac1{2 N} \sum_{i=1}^N \left\| \mW^L \vo_i^{(L-1)} - \tilde{\vx}_i \right\|^2 \\ &+ \frac{\alpha}{2\sqrt{\sum_{l=1}^L \#(\mW_1^l)}} \sum_{l=1}^L \left\| \mW^l - \mW_0^l \right\|^2, \numberthis \label{cfae}
\end{align*}
where $\| \cdot \|$ denotes the Frobenius norm and $\#(\mW_1^l)$ the number of rows of $\mW^l$. Let $\alpha$ be fixed to 1e-6 throughout; to simplify the notations, we define $\beta = \alpha/\sqrt{\sum_{l=1}^L \lvert\mW_1^l\rvert}$. The underlying idea in \eqref{cfae} is to average in both terms: in the first, the data fidelity (least-squares error, LSE) term, and in the second, the regularization term. Averaging the first term with $\frac1N$ implies that the term scales automatically by the size of the data subset, for instance, in minibatching, thereby providing an approximation of the entire LSE on a comparable scale. In the second term, because $\alpha$ is fixed, the inverse scaling constant $1/\sqrt{\sum_{l=1}^L \lvert\mW_1^l\rvert}$ 
balances the effect of the regularization compared to the data fidelity for networks with a different number of layers of different sizes. Because \eqref{cfae} will be minimized with local optimizers, we simply use the initial guesses $\{ \mW_0^l\}$ of the weight values in the second term to improve the local coercivity of \eqref{cfae} and to restrict the magnitude of the weights, thereby attempting to improve generalization \citep{gouk2021regularisation}. Because of the residual approximation, the random initialization of the weight matrices is generated from the uniform distribution ${\mathcal U} ([-0.1, 0.1])$. s

The gradient matrices $\nabla_{\mW^l} \cJ (\{\mW^l\}_{l=1}^L),\  l = L, \ldots, 1,$ for \eqref{cfae} are of the following form (see \cite{karkkainen2002mlp}):
\begin{equation}\label{FulOC}
\nabla_{\mW^l} \cJ (\{\mW^l\}) = \frac1N \sum_{i=1}^N \vd_i^l\,(\vo_i^{(l-1)})^T + \beta \left( \mW^l - \mW_0^l \right) ,
\end{equation}
where the layerwise error backpropagation reads as
\begin{eqnarray}
\displaystyle \vd_i^L &=& \displaystyle
\ve_i = \mW^L  \vo_i^{(L-1)} - \tilde{\vx}_i ,\label{diL}\\
\displaystyle \vd_i^l
&=& \displaystyle \Diag\{ (\cF)^{'} (\mW^l\,\vo_i^{(l-1)}) \}\,
(\mW^{(l+1)})^T\,\vd_i^{(l+1)} . \label{dil}
\end{eqnarray}
The use of different weights in the encoding\textemdash that is, in the transformation until layer $L/2$\textemdash and decoding, from layer $L/2$ to $L$, implies more flexibility in the residual autoencoder but also rougly doubles the amount of weights to be determined. Therefore, it is common to use the formal adjoint $(\mW^1)^T \cF ((\mW^2)^T \cF (\ldots (\mW^{L/2})^T))$ of the encoder as the decoder. Then, it is easy to see that the layerwise optimality conditions for $l = 1, \ldots, L/2$ read as
\begin{align*}
\grad_{\mW^l} \cJ = \frac1N \sum_{i=1}^N \vd_i^l\,(\vo_i^{(l-1)})^T + \vo_i^{(\tilde{l}-1)}\,\left(\vd_i^{\tilde{l}})^T \right) + \beta \left( \mW^l - \mW_0^l \right), \numberthis \label{SymOC}
\end{align*}
where $\tilde{l} = L - (l - 1)$. For convenience, we define $\tilde{L} = L/2$\textemdash that is, the number of layers to be optimized with the symmetric models.

We note that when the layerwise formulae above are used with vector-based optimizers, 
we always need to reshape operations to toggle between the weight matrices and a column vector of all weights.

\begin{rem}\label{rem:UseOfSeqAE}
Let us briefly summarize the use of the additive autoencoder for an unseen dataset after it has been estimated (and the corresponding data structures have been stored) for the training data through the three phases. First, data is normalized through mean subtraction and feature scaling into the same range $[-1, 1]$. Then, residuals according to formula \eqref{PCRes} are computed and this residual data is fed to the feedforward autoencoder. Again due to \eqref{PCRes}, the reduced, $m$-dimensional representation of new data is obtained as a sum of its PC projection and the output of the autoencoder's squeezing layer. Formula \eqref{PCRes} shows that the explicit formation of the residual data between linear and nonlinear representations can be replaced by setting $\widetilde{\mW}^1 = \mW^1 (\mI - \mU \mU^T)$ and using this as the first transformation layer of the autoencoder for the normalized, unseen data.
\end{rem}

\subsection{Layerwise pretraining}\label{subsec:Stacking}

\begin{figure*}[b!]
\centering
\includegraphics[width=10cm]{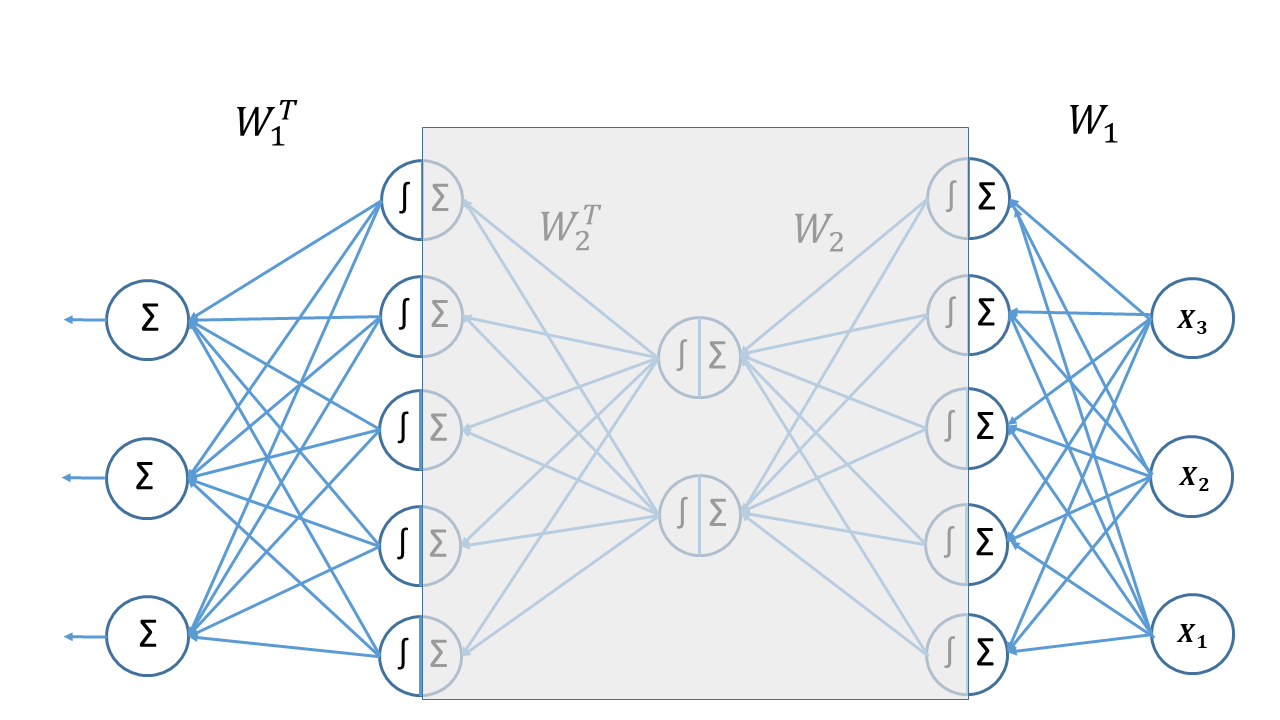}
\caption{Layerwise pretraining from heads to inner layers. The most outer layer is trained first and its residual is then fed as training data for the next hidden layer until all layers have been sequentially pretrained.}
\label{fig:PreTr}
\end{figure*}

The core idea of the proposed autoencoding model\textemdash additive combination of operators of different complexity\textemdash can be applied in the layerwise pretraining, that is, stacking of the nonlinear autoencoding part as well. We propose to use an identical network structure and learning paradigm for this purpose differently from, e.g., \citep{hinton2006reducing}. A similar idea appears with the deep residual networks (ResNets) in \cite{he2016deep}, where consecutive residuals are stacked together using layer skips\textemdash for example, over two or three layers with batch normalization. Moreover, in ResNets, the layer skips can introduce additional weight matrices to the deep network model. However, the layer-by-layer pretraining of the symmetric autoencoder, from the heads toward the inner layers, can be simply performed directly.

The approach is illustrated in Fig. \ref{fig:PreTr}. For three hidden layers with two unknown weight matrices, $\mW^1$ and $\mW^2$, we first estimate $\mW^1$ with the given data $\{ \tilde{\vx}_i \}$. Then, the output data of the estimated layer $\{\mW^1 \tilde{\vx}_i \}$ are used as the training data (the input and the desired output) for the second layer $\mW^2$. Thereafter, the entire network is fine-tuned by optimizing over both weight matrices. The process from the heads to the inner layers is naturally enlarged for a larger number of hidden layers. We could then also apply partial fine-tuning\textemdash for example, to fine-tune the three hidden layers during the process of constructing a five-hidden-layer network. However, according to our tests and similar to \cite{hinton2006reducing}, the layerwise pretraining suffices before fine-tuning the entire network. A special case of utilizing a simpler structure is the one-hidden-layer case: The symmetric model \nSymOne~ with 
one weight matrix if first optimizer to obtain $\mW^1$
and then used in the form $((\mW^1)^T, \mW^1)$ as the initial guess for optimizing the nonsymmetric model \nHidOne~with two weight matrices. Again such an approach could be generalized to multiple hidden-layer case for the nonsymmetric, deep autoencoding model.

\begin{rem}
As stated in the introduction, stacking attempts to mitigate the vanishing gradient problem, which may prevent the adaptation of the weights in deeper layers. We assessed the possibility of such a phenomenon by studying the relative changes in the weight matrix norms $(\big\lvert \| \mW_0^l \| - \| \mW_*^l \| \big\rvert / \| \mW_0^l \|)$ while fine-tuning the symmetric autoencoders with 3--7 layers (\nSymThree, \nSymFive, and \nSymSeven; see Section \ref{sec:Exprm}). The subscripts '0' and '$*$' refer to the initial and final weight matrix values, respectively. This study revealed that the relative changes in the weights in the deeper layers 
were not on a smaller numerical scale compared to the other layers. Apparently, the double role of the layers in the symmetric models as part of the encoder and the decoder, with the corresponding effect on the gradient as seen in formula \eqref{SymOC}, is also helpful in avoiding a vanishing gradient.
\end{rem}

\subsection{Determination of intrinsic dimension}\label{subsec:HidDim}

The basic procedure to determine the intrinsic dimension is to gradually increase the size of the squeezing layer and to seek a small value of the reconstruction error measuring autoencoding error, with a knee point \citep{thorndike1953belongs} indicating that the level of nondeterministic residual noise has been reached in autoencoding. Instead of the usual root mean squared error (RMSE), we apply the mean root squared error (MRSE) to compute the autoencoding error:
\begin{equation}\label{AE-RMSE}
e = \frac1{N} \sqrt{\sum_{i=1}^N \lvert\vx_i - \NV (\vx_i)\rvert^2} ,
\end{equation}
where $\{ \vx_i \}$ is assumed to be normalized and $\NV$ denotes the application of the autoencoder. This choice was made because in \cite{karkkainen2014cross}, MRSE correlated better with the independent validation error. In practice, the difference between the RMSE and the MRSE is only the scaling factor, $1/\sqrt{N}$ vs. $1/N$. After the linear PC trend estimation, the MRSE is obtained by using \eqref{AE-RMSE} for the residual data defined in formula \eqref{PCRes}. Note that the reconstruction error is a strict error measure and its use requires higher 
accuracy from autoencoding compared to other measures: with the Wine dataset in Fig.  \ref{fig:IntroIllustration}, the linear PCA needs all dimensions of the rotated coordinate axis for the reconstruction whereas already 10 principal components out of 13 would explain over 96\% of the data variance.

An example of determining the intrinsic dimension of the Glass dataset (see the next section) is presented in Fig. \ref{fig:GlassEx}. In the figure, the $x$-axis ``SqDim'' presents the squeezing dimension and the $y$-axis on the left the ``MRSE'' and on the right its change ``$\Delta$(MRSE)'' (i.e., backward difference) for the symmetric model with one hidden dimension (\nSymOne) and the corresponding nonsymmetric model \nHidOne. The intrinsic dimension of data is detected by first locating a sufficiently small change in the autoencoding error (the right plot). For this purpose, a user-defined threshold $\tau$ = 4e-3 is applied. The detected dimension 5 on the left, marked with a circle, is the dimension below the threshold on the right minus one. The intrinsic dimension 5 is also characterized by a clear knee point in the MRSE behavior.

\def\figurescale{5.9truecm}
\def\FigsCut{-0.2truecm}
\def\CaptCut{-0.0truecm}

\begin{figure*}[t!]
\centering
\includegraphics[width=\figurescale]{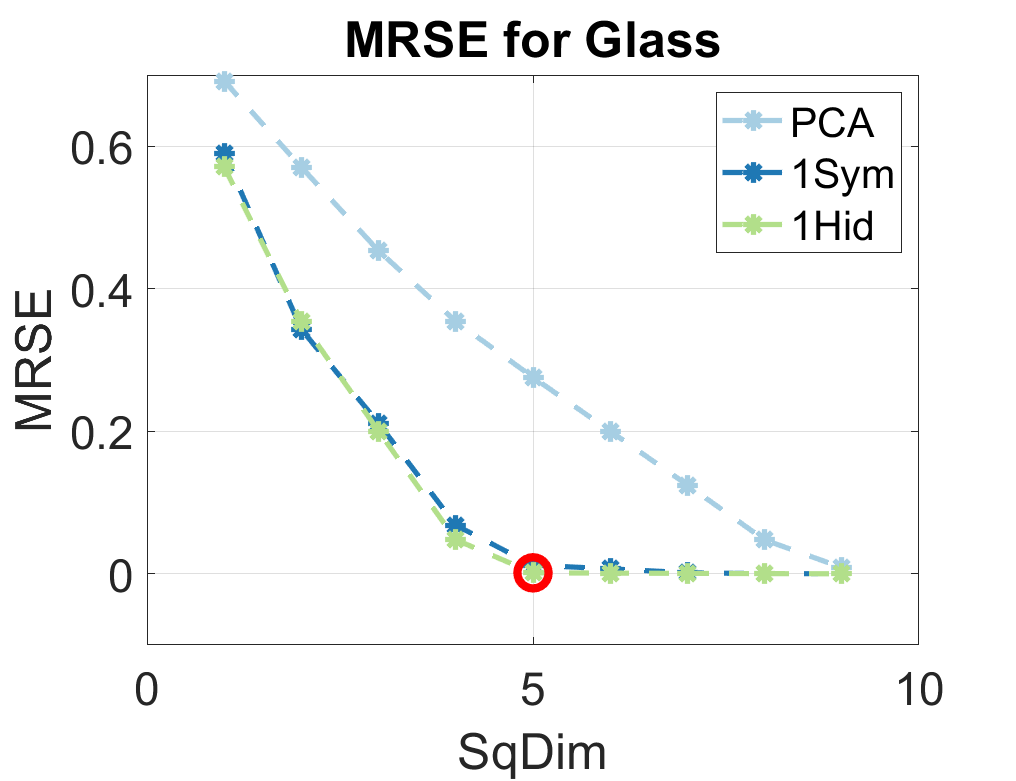}
\includegraphics[width=\figurescale]{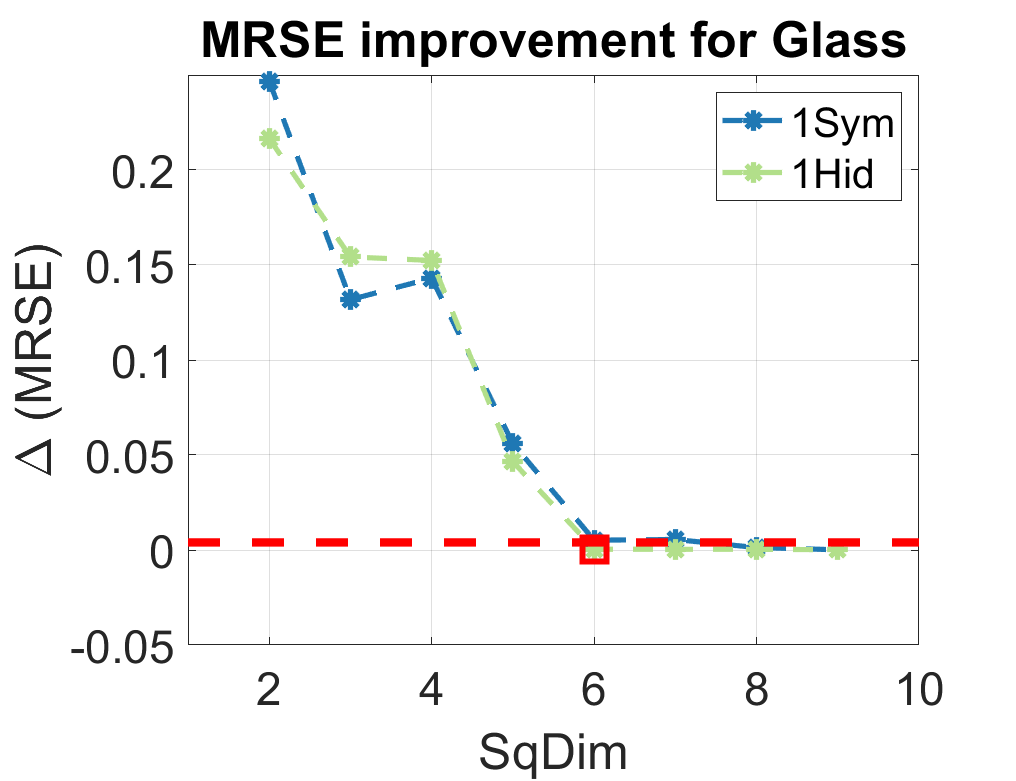}
\hspace*{\FigsCut}
\vspace*{\CaptCut}
\caption{Identification of the intrinsic dimension for the Glass dataset. The hidden dimension (plus one) on the left is captured by the sufficiently small error improvement on the right.}
\label{fig:GlassEx}
\end{figure*}

\section{Results}\label{sec:Exprm}

The main focus of the computational experiments, which are fully reported in the Supplementary Information (SI), was to investigate the ability of the proposed additive autoencoder model to represent a dataset in a lower-dimensional space. Therefore, we confined ourselves to the use of Matlab as the platform (mimicking the experiments in \cite{hinton2006reducing}) to have full control over the realization of the methods in order to study the effects of different parameters and configurations. 
Reference implementation of the proposed method and its basic testing is available in GitHub\footnote{\url{https://github.com/TommiKark/AdditiveAutoencoder}}.

We apply and compare the following set of techniques to approximate the nonlinear residual of the autoencoder, after normalization and identification of the linear trend: \nHidOne~(model with one hidden layer and separate weight matrices for the encoder and the decoder), \nSymOne~(symmetric model with one hidden layer and a shared weight matrix), \nSymThree~(three-hidden-layer symmetric model with two shared weight matrices), \nSymFive, and \nSymSeven. To systematically increase the flexibility and the approximation capability of the deeper models, the sizes of the layers for $l = \tilde{L}, \ldots, 1$ are given below, where $n_{\tilde{L}}$ is the size of the squeezing layer:
\begin{description}
\item{\nSymThree:} $n_{\tilde{L}}$ -- $2 n_{\tilde{L}}$ -- $n$,
\item{\nSymFive:} $n_{\tilde{L}}$ -- $2 n_{\tilde{L}}$ -- $4 n_{\tilde{L}}$ -- $n$,
\item{\nSymSeven:} $n_{\tilde{L}}$ -- $2 n_{\tilde{L}}$ -- $3 n_{\tilde{L}}$ -- $4 n_{\tilde{L}}$  -- $n$.
\end{description}
Note that for $n_{\tilde{L}} > n/2$ the size of the second layer and, therefore, the dimension of the first intermediate representation, is larger than the input dimension for all these models.

\subsection{Identification of the intrinsic dimension}\label{subsec:DimExp}

The first purpose of the experiments was to search for the intrinsic dimension of a dataset via autoencoding. This was done using the shallow models \nSymOne~and \nHidOne. The optimization settings and visualization of all results are given in the online SM.

The experiments were carried out for two groups of datasets, one with small-dimension data (less than 100 features) and the other with large-dimension data (up to 1024 features). The datasets were obtained from the UCI repository \citep{Dua:2019}, except the FashMNIST, which was downloaded  from GitHub\footnote{\url{https://github.com/zalandoresearch/fashion-mnist}}. For most of the datasets, only the training data was used; however, with Madelon, the given training, testing, and validation datasets were combined. The datasets do not contain missing values. The constant features were identified and eliminated according to whether the difference between the maximum and minimum values of a feature was less than $\sqrt{MEps}$, where $MEps$ denotes machine epsilon (this is classically used numerical proxy of zero, see \cite[p.~12]{dennis1996numerical}). Because of this preprocessing, the number of features $n$ in Tables \ref{DataTableSmall} and \ref{DataTableLarge} is not necessarily the same as that in the UCI repository.

\begin{table}[t!]
\centering
\caption{Results of the identification of the intrinsic dimension for small-dimension datasets. The intrinsic dimensions were identified with the reduction rates varying between 0.41--0.54. The SteelPlates and COIL2000 (with the most discrete feature profile) have the best reduction rate. The residual errors are between 1.1e-2--4.3e-4.}\label{DataTableSmall}
\begin{tabular}{llcclll}
\toprule
Dataset & $N$ & $n$ & ID & Red & MRSE & FeatProf (\%) \\
\midrule
Glass & 214 & 10 & 5 & 0.50 & 1.3e-3 & 10-40-50-0 \\
Wine & 178 & 13 & 7 & 0.54 & 1.2e-3 & 0-46-54-0 \\
Letter & 20 000 & 16 & 8 & 0.50 & 9.4e-4 & 0-100-0-0 \\
SML2010 & 2 763 & 17 & 9 & 0.53 & 5.4e-4 & 0-12-18-71 \\
FrogMFCC & 7 195 & 22 & 11 & 0.50 & 1.1e-3 & 0-0-5-95 \\
SteelPlates & 1 941 & 27 & 11 & 0.41 & 4.3e-3 & 11-11-56-22 \\
BreastCancerW & 569 & 30 & 14 & 0.47 & 6.9e-3 & 0-0-100-0 \\
Ionosphere & 351 & 33 & 17 & 0.52 & 1.9e-3 & 3-0-97-0 \\
SatImage & 6 435 & 36 & 18 & 0.50 & 4.3e-4 & 0-75-25-0 \\
SuperCond & 21 263 & 82 & 37 & 0.45 & 1.1e-2 & 2-1-12-84 \\
COIL2000 & 5 822 & 85 & 35 & 0.41 & 2.8e-2 & 99-1-0-0 \\
\bottomrule
\end{tabular}
\end{table}

\begin{table}[t!]
\centering
\caption{Results of the identification of the intrinsic dimension for large-dimension datasets. The intrinsic dimensions were identified with the reduction rates varying between 0.39--0.55. The HumActRec dataset with a continuous feature profile has the best reduction rate. The residual errors are between 8.0e-2--2.9e-3.}\label{DataTableLarge}
\begin{tabular}{llcclll}
\toprule
Dataset & $N$ & $n$ & ID & Red & MRSE & FeatProf (\%) \\ 
\midrule
USPS & 9 298 & 256 & 130 & 0.51 & 2.9e-3 & 0-0-0-100 \\
BlogPosts & 52 397 & 277 & 130 & 0.47 & 3.9e-3 & 79-8-12-1 \\
CTSlices & 53 500 & 379 & 180 & 0.47 & 6.3e-2 & 8-4-10-78 \\
UJIIndoor & 19 937 & 473 & 200 & 0.42 & 7.0e-2 & 26-73-1-0 \\
Madelon & 4 400 & 500 & 250 & 0.50 & 7.9e-2 & 2-31-67-0 \\
HumActRec & 7 351 & 561 & 220 & 0.39 & 8.0e-2 & 0-2-0-98 \\
Isolet & 7 797 & 617 & 310 & 0.50 & 9.4e-3 & 1-6-14-80 \\
MNIST & 60 000 & 717 & 350 & 0.49 & 9.3e-3 & 9-14-77-0 \\
FashMNIST & 60 000 & 784 & 380 & 0.48 & 5.0e-2 & 0-2-98-0 \\
COIL100 & 7 200 & 1 024 & 560 & 0.55 & 6.4e-3 & 0-11-89-0 \\
\bottomrule
\end{tabular}
\end{table}

During the search, the squeezing dimension for the small-dimension datasets began from one and was incremented one by one up to $n-1$. For the large-dimension cases, we began from 10 and used increments of $10$ until the maximum squeezing dimension $\lfloor 0.6\times n \rfloor$ was reached (cf. the ``Red'' values in Tables \ref{DataTableSmall} and \ref{DataTableLarge}). The experiments were run with Matlab on a Laptop with 2.3GHz Intel i7 processor and 64 GB RAM and on a server with a Xeon E5-2690 v4 CPU and 384 GB of memory. 

In Tables \ref{DataTableSmall} and \ref{DataTableLarge}, we present the name of the dataset, the number of observations $N$, the number of features $n$, and the detected intrinsic dimension ID. The autoencoding error trajectories and thresholdings are illustrated for all datasets in the online SM. The detection threshold for small-dimension datasets was fixed to $\tau$ = 4e-3. The reduction rate ID$/n$ for the intrinsic data dimension is reported in the Red column, and the autoencoding error of \nHidOne~ for ID according to \eqref{AE-RMSE} is included in the MRSE column. There is no averaging over the data dimension $n$ in \eqref{AE-RMSE}, so that for higher-dimension datasets this error is expected to remain larger. This was probably one of the reasons why, for large-dimension datasets, we needed to use two values of the threshold $\tau$ (based on visual inspection; see the zoomed illustrations in the SM): 3e-3 for USPS, BlogPosts, HumActRec, MNIST, and COIL100, and 3e-2 for the remaining five datasets.

\def\figurescale{5.8truecm}
\def\FigsCut{-0.1truecm}

\begin{figure*}[b!]
\centering
\hspace*{\FigsCut}
\includegraphics[width=\figurescale]{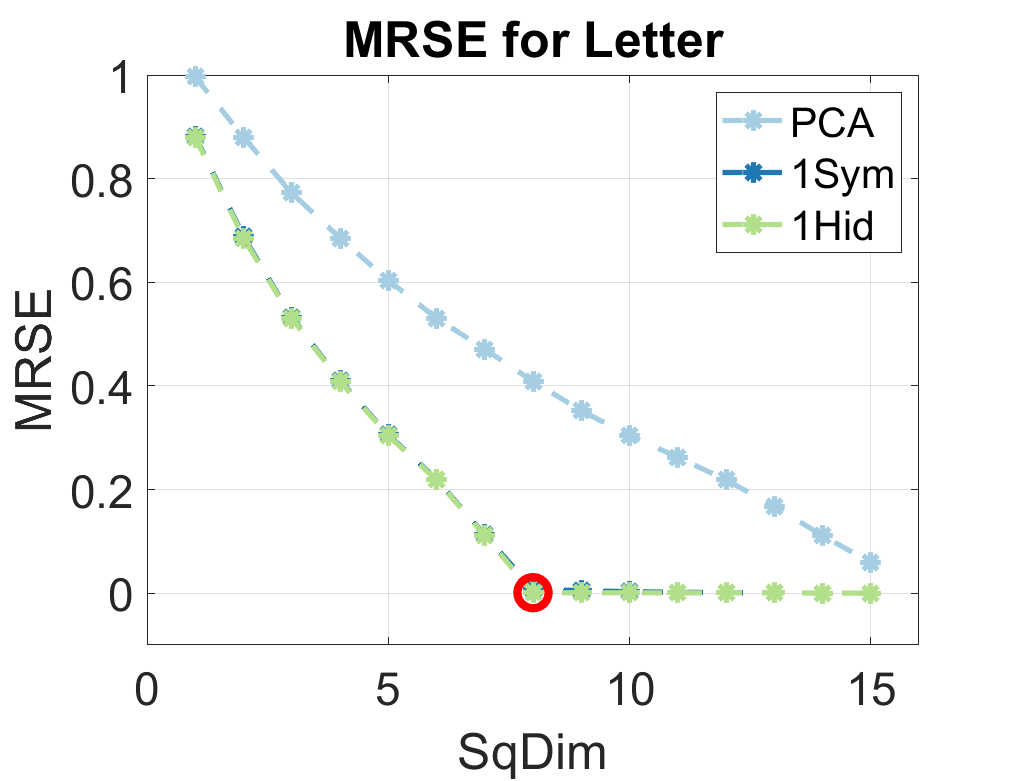}
\hspace*{\FigsCut}
\includegraphics[width=\figurescale]{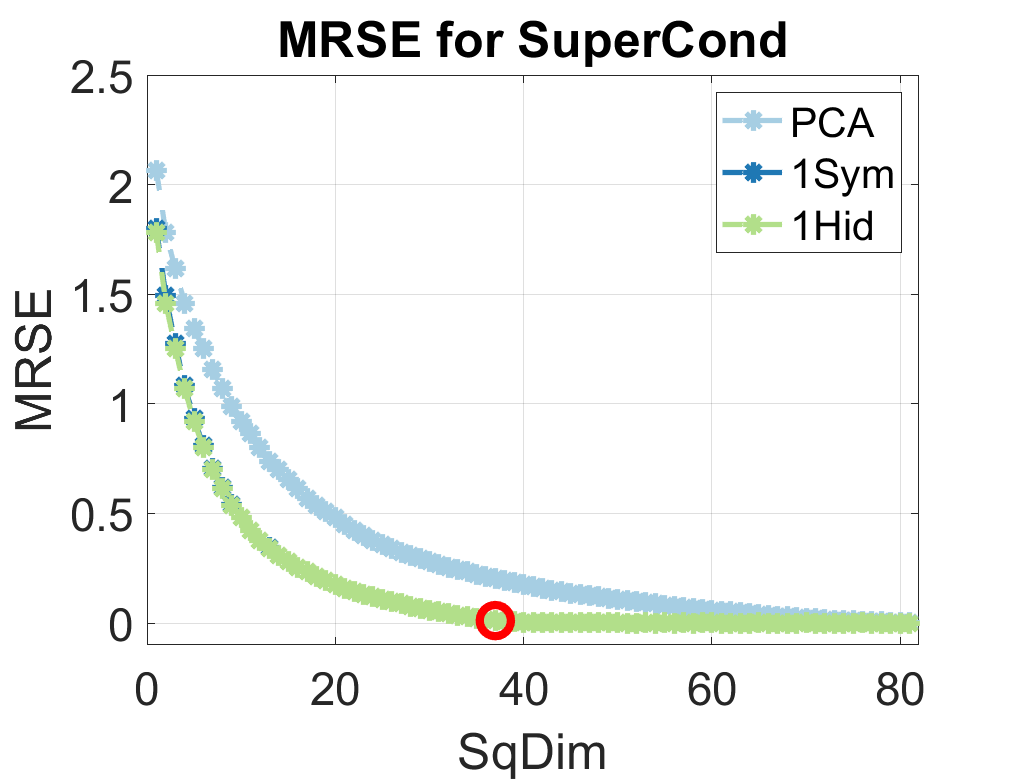}
\caption{Identification of the intrinsic dimension for the Letter dataset and SuperCond dataset. \textbf{Left}: Clearly identified knee-point in ID = 8 for Letter. \textbf{Right}: Gradual decrease of MRSE with ID = 37 for SuperCond.}
\label{fig:LetterSuperCondHidDims}
\end{figure*}

\begin{figure*}[hbt!]
\centering
\hspace*{\FigsCut}
\includegraphics[width=\figurescale]{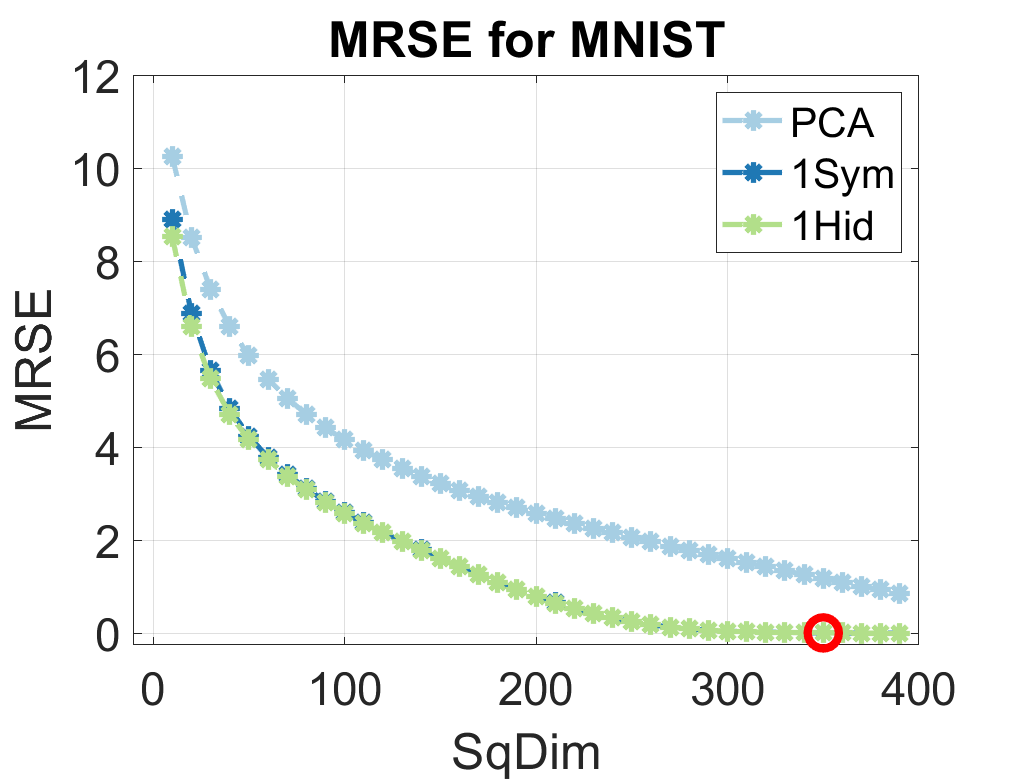}
\includegraphics[width=\figurescale]{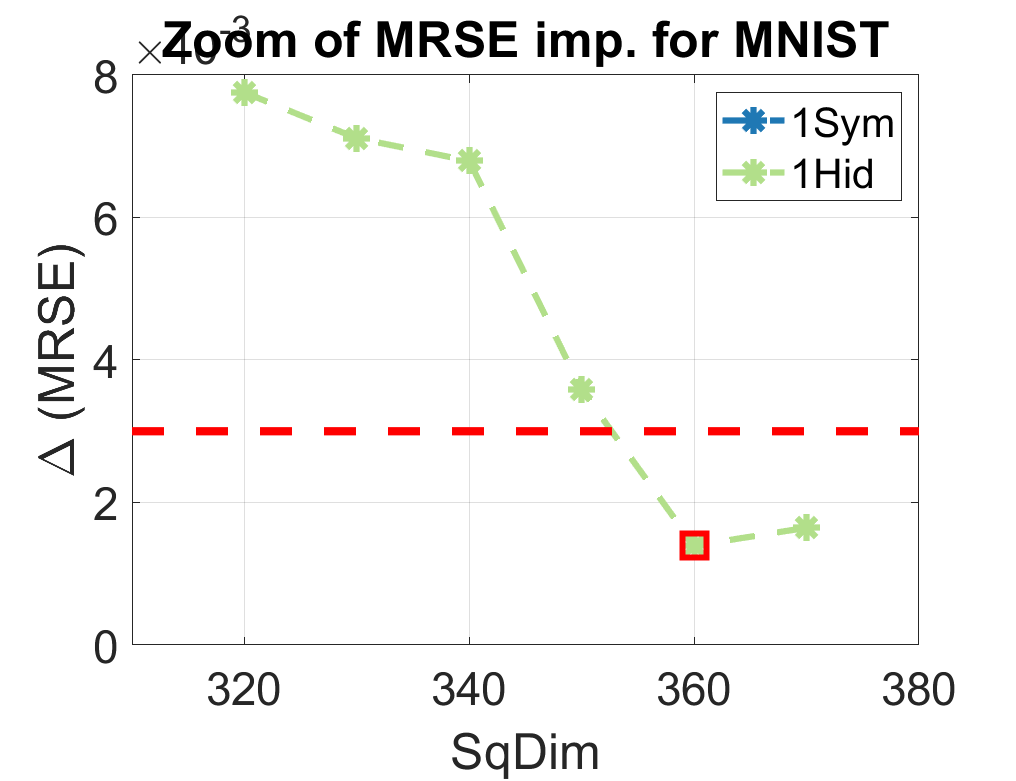}
\caption{Identification of the intrinsic dimension for the MNIST dataset. The hidden dimension (plus one) on the left is captured by the sufficiently small error improvement on the right. \textbf{Left}: Gradual decrease of MRSE with ID = 350 for MNIST. \textbf{Right}: Zoom of MRSE improvement with the threshold $\tau$ = 3e-3 confirms the detection.}
\label{fig:MNISTHidDim}
\end{figure*}

\begin{figure*}[hbt!]
\centering
\hspace*{\FigsCut}
\includegraphics[width=\figurescale]{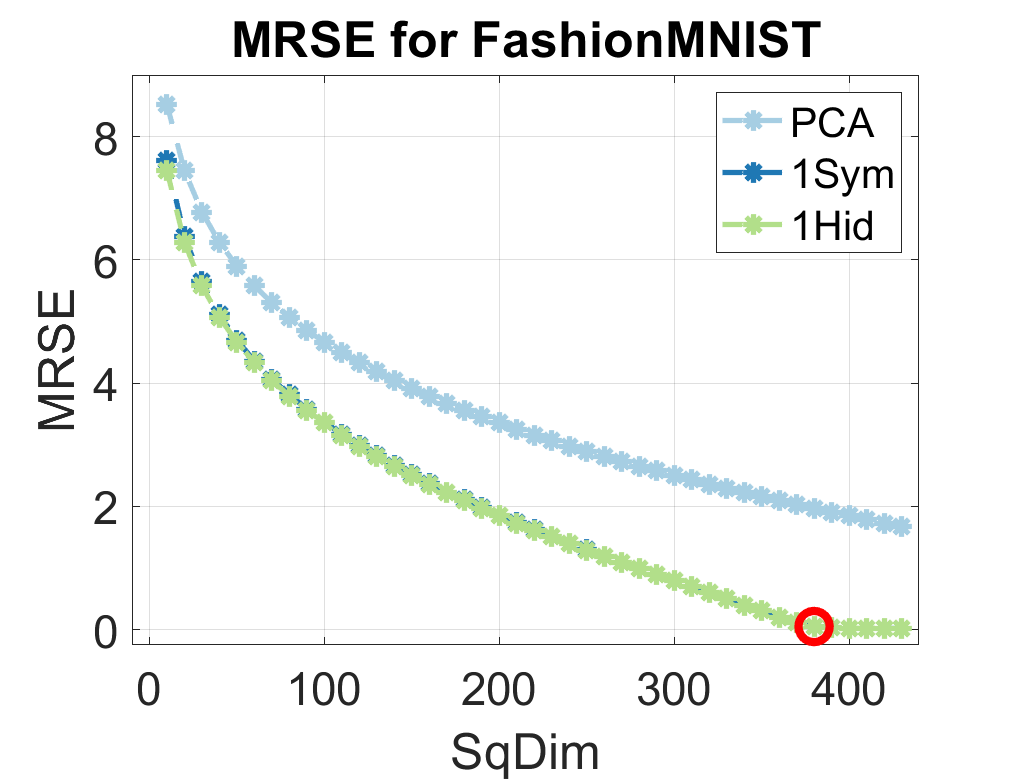}
\includegraphics[width=\figurescale]{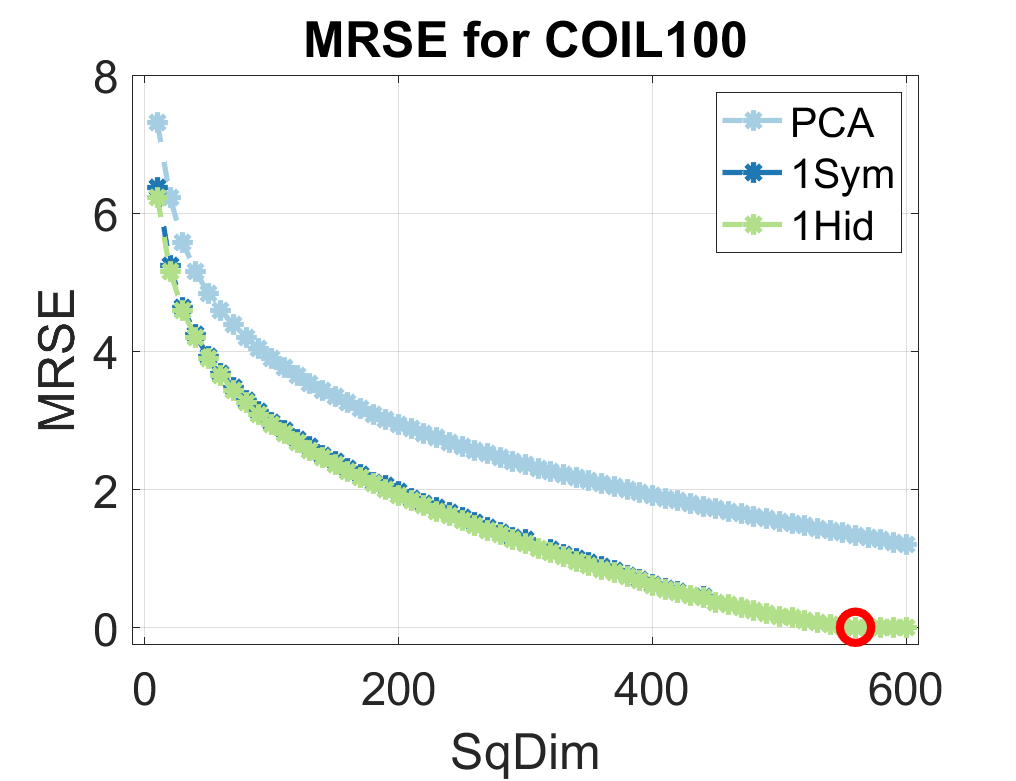}
\caption{Identification of the intrinsic dimension for the FashionMNISTS dataset and COIL100 dataset. \textbf{Left}: Clear knee-point of MRSE with ID = 380 for FashMNIST. \textbf{Right}: More gradual decrease of MRSE with ID = 560 for COIL100.}
\label{fig:FashionMNISTCOIL100HidDim}
\end{figure*}

For the analysis, we also included a depiction of how discrete or continuous the set of features for a dataset is. We categorized the features into four groups based on the number of unique values (UV) each feature has: C1 = \{UV $\le$ 10\}, C2 = \{10 $<$ UV $\le$ 100\}, C3 = \{100 $<$ UV $\le$ 1000\}, and C4 = \{1000 $<$ UV\}. The FeatProf column in Tables \ref{DataTableSmall} and \ref{DataTableLarge} presents the proportions of C1--C4 in percentages.

\subsubsection*{Conclusions}

Examples of the ID detection are given in Figs. \ref{fig:GlassEx} (Glass), \ref{fig:LetterSuperCondHidDims} (Letter on the left, SuperCond on the right), \ref{fig:MNISTHidDim} (MNIST), and \ref{fig:FashionMNISTCOIL100HidDim} (FashMNIST on the left, COIL100 on the right). Identifications in the first two cases and for FashMNIST are characterized by clear knee-points in IDs. For SuperCond, with gradual decrease of the MRSE, determination of ID is based on the mutual threshold value $\tau$ = 4e-3 of small-dimension datasets. Also MNIST has such a behavior and the zoom in Fig. \ref{fig:MNISTHidDim} (right) illustrates the detection decision with $\tau$ = 3e-3.

Overall, the intrinsic dimensions were successfully identified for all tested datasets. The use of a feedforward network to approximate the nonlinear residual notably decreased the autoencoding error of the linear PCA. The overall transformation summarizing the essential behavior of data roughly halved the original dimension: The mean reduction rate over the 21 datasets was 0.48.

The reduction rate was independent of the form of the features\textemdash that is, the best reduction rates for small-dimension datasets were obtained with the very categorical COIL2000 and primarily continuous SteelPlates datasets. However, the best reduction rate, 0.39, was obtained for HumActRec, which is characterized by a continuous feature profile. This indicates that we may obtain smaller reduction rates with more continuous sets of features.

\subsection{Comparison of shallow and deep models}\label{subsec:OptExp3}

\begin{table*}[b!]
\begin{center}
\caption{Efficiencies of symmetric models for small-dimension datasets.}\label{tab:RelErrsSmallDatasets}
{
\resizebox{\linewidth}{!}{
\begin{tabular}{p{2.4truecm} | p{0.6truecm} p{1.55truecm} | p{0.6truecm} p{1.55truecm} | p{0.6truecm} p{1.55truecm} | p{0.6truecm} p{1.55truecm} }
\toprule
  & \multicolumn{2}{c|}{\nSymOne} & \multicolumn{2}{c|}{\nSymThree} & \multicolumn{2}{c|}{\nSymFive} & \multicolumn{2}{c}{\nSymSeven} \\ 
  Dataset & mean & max (dim) & mean & max (dim) & mean & max (dim) & mean & max (dim) \\ 
\midrule
 Glass & 0.91 & 1.03 (2) & 0.91 & 1.10 (3) & 1.18 & 1.31 (3) & 1.15 & 1.40 (3) \\
 Wine & 0.97 & 0.99 (1) & 1.07 & 1.22 (6) & 1.25 & 1.59 (6) & 1.36 & 1.75 (6) \\
 Letter & 1.00 & 1.00 (1) & 1.03 & 1.06 (6) & 1.10 & 1.14 (6) & 1.11 & 1.15 (6) \\
 SML2010 & 0.94 & 0.99 (1) & 0.95 & 1.07 (5) & 1.12 & 1.23 (3) & 1.15 & 1.32 (3) \\
 FrogMFCCs & 0.99 & 1.00 (7) & 1.04 & 1.17 (8) & 1.10 & 1.23 (8) & 1.11 & 1.23 (8) \\ 
 SteelPlates & 0.94 & 0.99 (4) & 0.94 & 1.09 (5) & 1.04 & 1.22 (5) & 1.04 & 1.27 (5) \\
 BreastCancerW & 0.98 & 0.99 (11) & 1.10 & 1.29 (13) & 1.22 & 1.42 (12) & 1.24 & 1.41 (11) \\
 Ionosphere & 0.91 & 0.97 (3) & 1.04 & 1.26 (15) & 1.74 & 3.19 (14) & 2.07 & 4.06 (15) \\
 Satimage & 0.99 & 1.00 (10) & 1.02 & 1.07 (17) & 1.05 & 1.09 (17) & 1.06 & 1.12 (1) \\
 SuperCond & 1.00 & 1.00 (30) & 1.10 & 1.18 (33) & 1.20 & 1.30 (29) & 1.23 & 1.34 (26) \\
 COIL2000 & 0.99 & 1.02 (16) & 1.24 & 1.85 (32) & 1.49 & 2.89 (29) & 1.48 & 2.62 (30) \\ 
\bottomrule
\end{tabular}
}}
\end{center}
\end{table*}

\begin{table*}[t!]
\begin{center}
\caption{Efficiencies of symmetric models for large-dimension datasets.}\label{tab:RelErrsLargeDatasets}
{
\resizebox{\linewidth}{!}{
\begin{tabular}{p{2.4truecm} | p{0.6truecm} p{1.65truecm} | p{0.6truecm} p{1.85truecm} | p{0.6truecm} p{1.75truecm} | p{0.6truecm} p{1.7truecm} }
\toprule
  & \multicolumn{2}{c|}{\nSymOne} & \multicolumn{2}{c|}{\nSymThree} & \multicolumn{2}{c|}{\nSymFive} & \multicolumn{2}{c}{\nSymSeven} \\ 
  Dataset & mean & max (dim) & mean & max (dim) & mean & max (dim) & mean & max (dim) \\
\midrule
 USPS & 0.99 & 1.00 (90) & 1.08 & 1.16 (90) & 1.13 & 1.21 (70) & 1.14 & 1.23 (60) \\
 BlogPosts & 0.95 & 1.00 (110) & 1.18 & 1.39 (70) & 1.28 & 1.56 (80) & 1.23 & 1.53 (70) \\
 CTSlices & 0.99 & 1.00 (170) & 1.17 & 1.51 (150) & 1.30 & 1.76 (150) & 1.32 & 1.74 (150) \\
 UJIIndoor & 0.99 & 0.99 (60) & 1.69 & 2.46 (150) & 2.15 & 3.11 (130) & 2.24 & 3.51 (130) \\
 Madelon & 0.97 & 1.00 (30) & 1.38 & 3.74 (240) & 2.29 & 5.77 (220) & 3.01 & 8.02 (220) \\ 
 HumActRec & 0.99 & 1.00 (160) & 1.15 & 1.31 (160) & 1.22 & 1.40 (130) & 1.24 & 1.40 (130) \\
 Isolet & 0.99 & 1.00 (290) & 1.22 & 2.16 (290) & 1.44 & 2.89 (270) & 1.55 & 2.65 (270) \\
 MNIST & 0.99 & 1.00 (200) & 1.32 & 1.99 (260) & 1.42 & 2.20 (250) & 1.41 & 2.25 (240) \\
 FashMNIST & 0.99 & 1.00 (320) & 1.15 & 1.63 (370) & 1.22 & 1.59 (350) & 1.23 & 1.53 (350) \\
  COIL100 & 0.98 & 1.00 (310) & 2.18 & 11.19 (520) & 1.96 & 8.51 (530) & 1.79 & 2.63 (430) \\
 COIL100-Min & 0.98 & 1.00 (310) & 1.75 & 8.05 (510) & 1.79 & 4.22 (510) & 1.87 & 2.63 (430) \\
 \bottomrule
\end{tabular}
}}
\end{center}
\end{table*}

\begin{figure*}[t!]
\centering
\hspace*{\FigsCut}
\includegraphics[width=\figurescale]{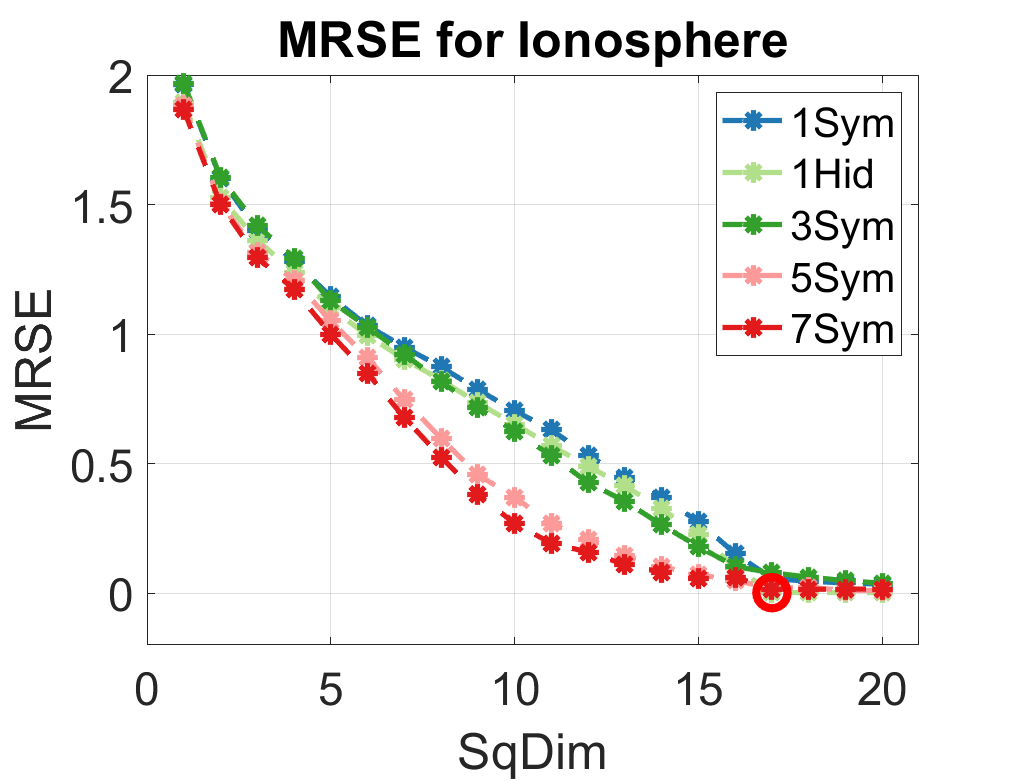}
\includegraphics[width=\figurescale]{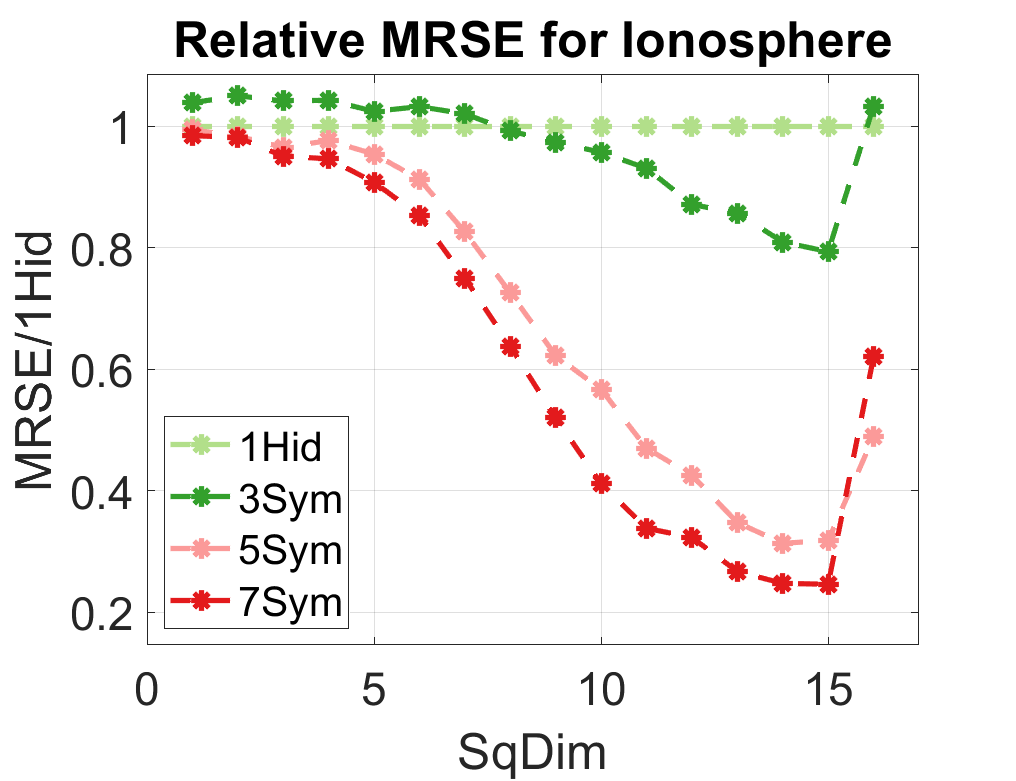}
\caption{\textbf{Left}: Behavior of the MRSE with all residual models for Ionosphere. \textbf{Right}: Relative performance of the models with respect to \nHidOne. During the search of the ID the deeper models show clear improvement but the detected ID is the same for all models.} 
\label{fig:IonosphereModelComp}
\end{figure*}

\begin{figure*}[t!]
\centering
\hspace*{\FigsCut}
\includegraphics[width=\figurescale]{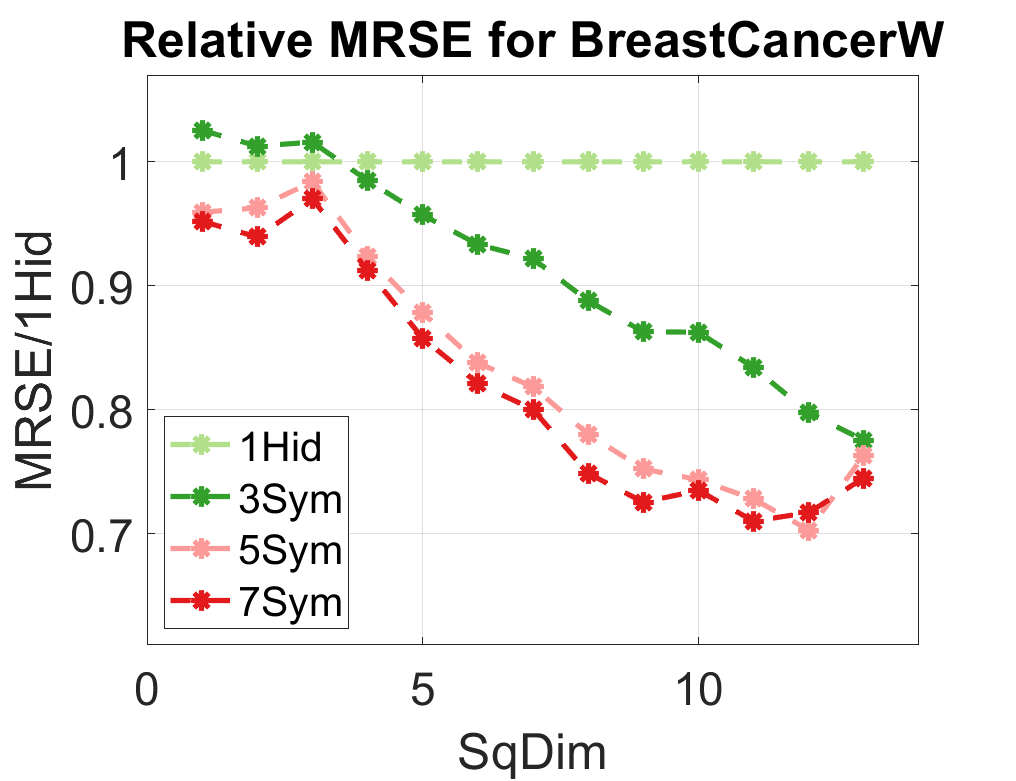}
\includegraphics[width=\figurescale]{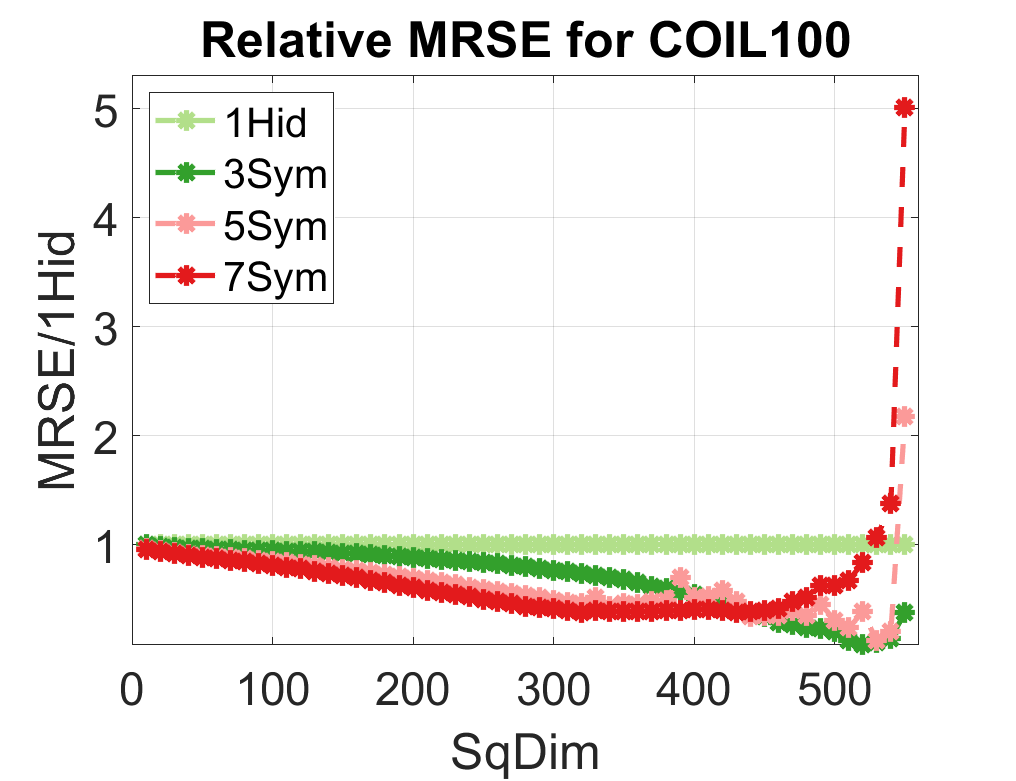}
\caption{\textbf{Left}: Relative performance of the models for BreastCancerW. \textbf{Right}: Relative performance of the models for COIL100. The deeper models show clear improvement over the shallow ones but the detected ID stays the same and near ID the improved efficiency may be completely lost.}  
\label{fig:BreastCandCOIL100H1Rel}
\end{figure*}

The second aim of the experiments was to examine whether deeper network structures and deep learning techniques (the network structure and optimization of the weights) can improve the identification of the intrinsic dimension and the data restoration ability of the additive autoencoder. This aim is pursued as follows: Here, we compare shallow and deep networks in cases where fine-tuning is performed using a classical optimization approach, i.e., using the L-BFGS optimizer with the complete dataset. 
In the SI, we report the results of using different minibatch-based approaches. Also detailed depictions of the parameter choices and visualization of the results for all datasets are included there.

In addition to visual assessment, we performed a quantitative comparison between the deep and shallow models. First, the MRSE values of all models were divided with the corresponding value of the \nHidOne~model's error. This was done for the squeezing dimensions from the first until next to last of ID, to cover the essential search phase of the intrinsic dimension. To exemplify relative performance, if the MRSE value of a model divided by the \nHidOne's value for a particular squeezing dimension would be 0.5, then such a model would have half the error level and, conversely, twice the efficiency compared to \nHidOne. Therefore, the model's efficiency is defined as the reciprocal of relative performance.

The relative performances are illustrated in Figs. \ref{fig:IonosphereModelComp} and \ref{fig:BreastCandCOIL100H1Rel}. Descriptive statistics of the efficiencies of symmetric models are given in Tables \ref{tab:RelErrsSmallDatasets} and \ref{tab:RelErrsLargeDatasets}. In each cell there, both the mean efficiency and the maximal efficiency are provided. The latter includes, in parentheses, the squeezing dimension where it was encountered.

\subsubsection*{Conclusions}

During the early phases of searching the intrinsic dimension, deep networks provide smaller autoecoding errors compared to the shallow models. However, as exemplified in Fig. \ref{fig:IonosphereModelComp} (left) and is evident from all illustrations in the SI, MRSE in ID is not better for deeper models compared to \nHidOne. Therefore, the use of a deeper model would not change the ID values and, in fact, fluctuation of the error compared to \nHidOne~may hinder the detection of a knee-point and negatively affect the simple thresholding.

Usually, the mean efficiency of the two deepest models, \nSymFive~and \nSymSeven, is better than that of \nSymOne~or \nSymThree, but for many datasets, there is only a slight improvement. Overall, the plots for the relative efficiencies between different symmetric models and the quantitative trends in the rows of Tables \ref{tab:RelErrsSmallDatasets} and \ref{tab:RelErrsLargeDatasets} include varying patterns.
The mean efficiency is highest for UJIIndoor, Ionosphere, Madelon, and COIL100, where the last three datasets are characterized by high data dimension/number of observations, $n/N$, ratio (0.09, 0.11, and 0.08, respectively). The following are the grand means of the mean efficiencies over all 21 datasets for the symmetric models: \nSymOne~0.97, \nSymThree~1.19, \nSymFive~1.38, and \nSymSeven~1.44. This concludes that deeper models improve the reduction of the autoencoding error during the search of ID. However, the speed of improvement decreases as a function of the number of layers.

\begin{figure*}[t!]
\centering
\hspace*{\FigsCut}
\includegraphics[width=\figurescale]{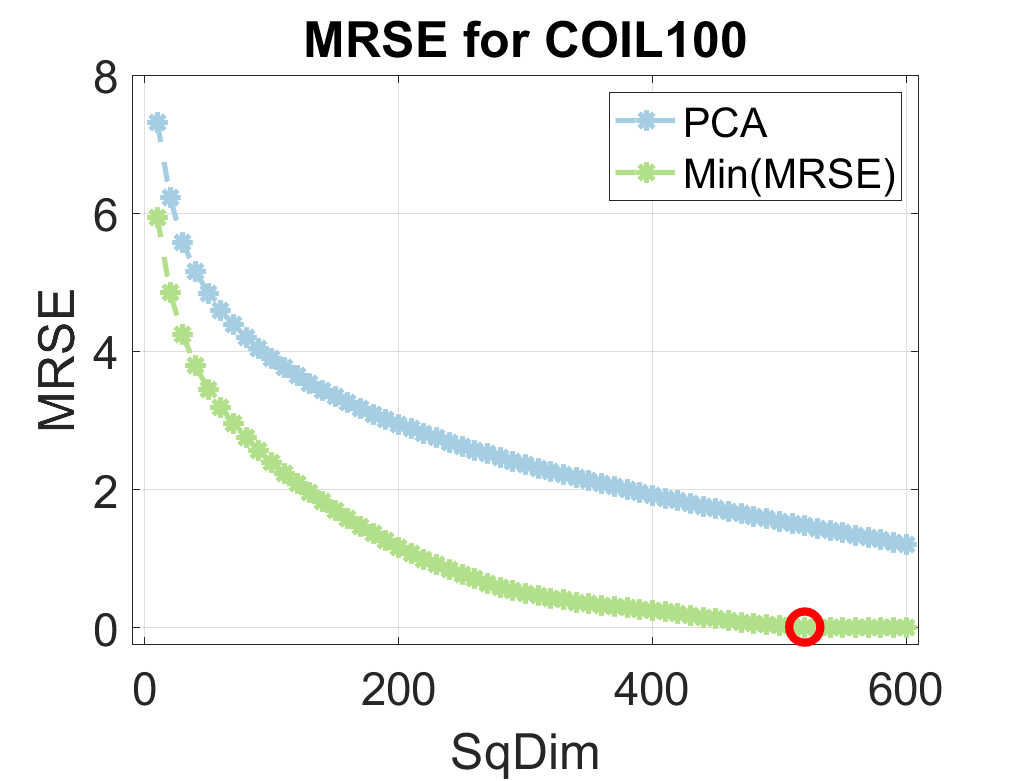}
\includegraphics[width=\figurescale]{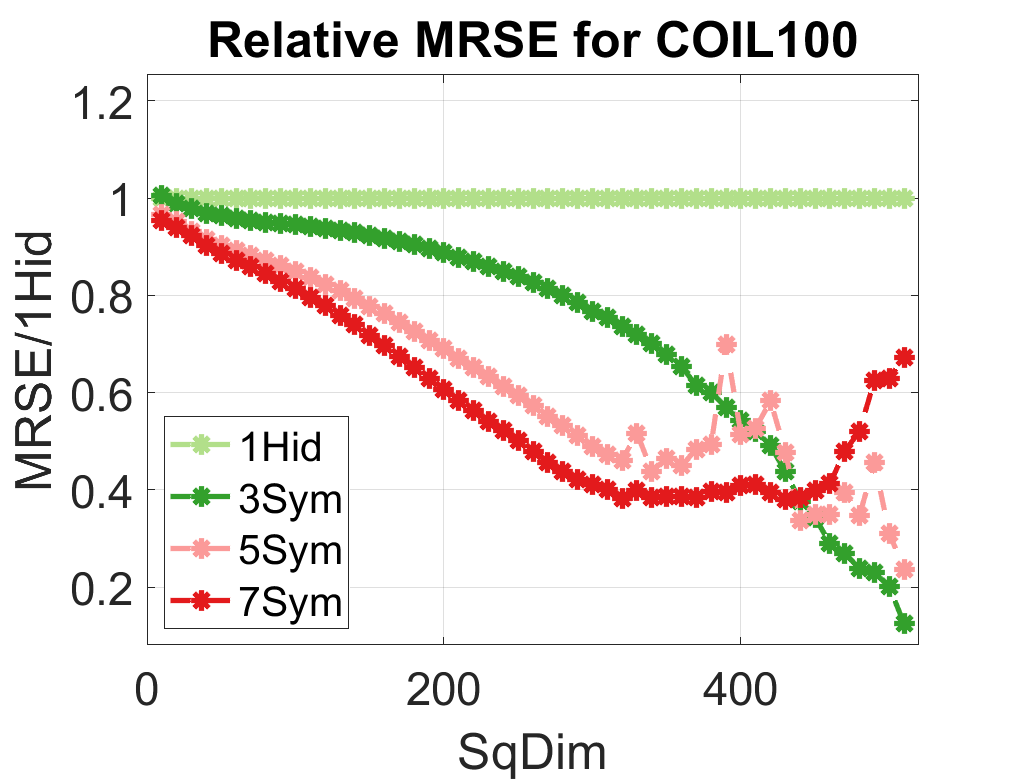}
\caption{\textbf{Left}: Minimum autoencoding error of the all models for COIL100. \textbf{Right}: Reduced relative performance of the models for COIL100. For COIL100, use of minimum autoencoding error to identify ID yielded more reasonable results.}
\label{fig:COIL100MinIDRes}
\end{figure*}

Actually, close to the intrinsic dimension, the benefits of deeper models may be completely lost. This is illustrated in Fig. \ref{fig:BreastCandCOIL100H1Rel} (right) and in Table \ref{tab:RelErrsLargeDatasets} for COIL100 (see also, for example, plots of BlogPosts and MNIST in the SI). The reason for such a behavior with COIL100 is the value of ID, 560, which was obtained with the smaller threshold $\tau$ = 3e-3 for large-dimension datasets. Therefore, with COIL100, we also tested an alternative approach to the identification of ID, where we apply the same thresholding technique (and the same $\tau$) to the minimum autoencoding error of the all models. This error plot, the identified ID = 520 (for which the reduction rate would be 0.51), and the corresponding reduced set of relative efficiencies are illustrated in Figure \ref{fig:COIL100MinIDRes}. The summary of the efficiencies for this modified way to identify ID are given in the last line ``COIL100-Min'' in Table \ref{tab:RelErrsLargeDatasets}. It can be concluded that for the largest dimensional dataset COIL100, the use of the minimum autoencoding error of the models yielded more reasonable results.

We used a fixed pattern for the sizes of the hidden layers in deeper models: 2-3-4 times the squeezing dimension for \nSymSeven, the first and last of these coefficients for \nSymFive, and the first one for \nSymThree. As reported in Section \ref{subsec:DimExp}, the mean reduction rate over the 21 datasets was close to 0.5. Therefore, one may wonder whether this behavior is due to the fact that from this case onwards all the hidden dimensions are larger than the number of features so that a kind of nonlinear kernel trick occurs. In other words, would a different pattern of the hidden dimensions change the results and conclusions here? This consideration was tested by considering a 3-5-7 pattern providing much more flexibility for the nonlinear operator compared to the used pattern. These tests are not reported as a whole, because the clearly identified trend of the results is readily exemplified in Fig. \ref{fig:BreastCancerHidPatComp}: Increase of the sizes of the hidden layers slightly improve the reduction rate during early phase of the search but does not change the value of the ID.

\def\figurescalemod{5.9truecm}
\def\figurescalemodmod{5.6truecm}

\begin{figure*}[t!]
\centering
\hspace*{\FigsCut}
\includegraphics[width=\figurescalemod]{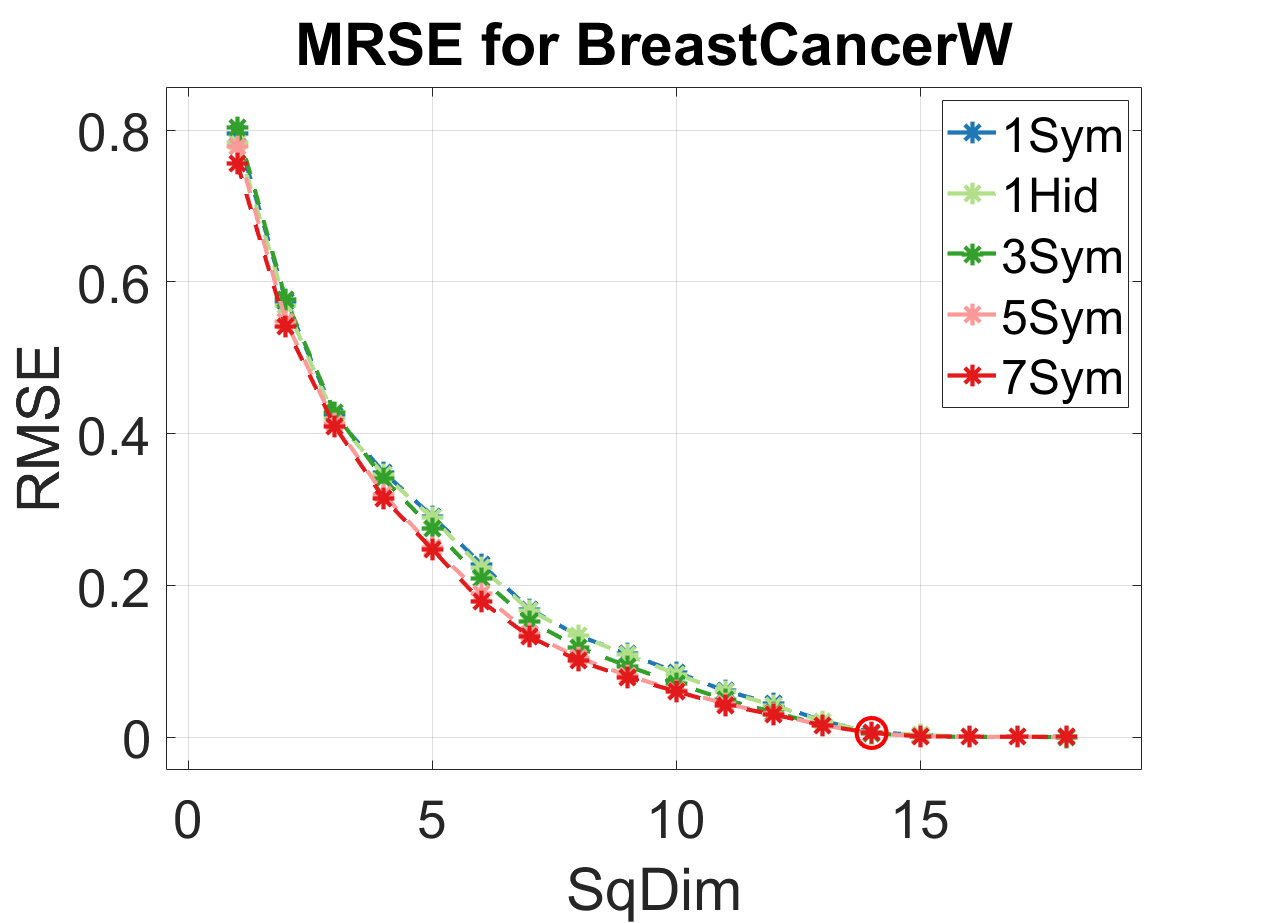}
\hspace*{0.1truecm}
\includegraphics[width=\figurescalemodmod]{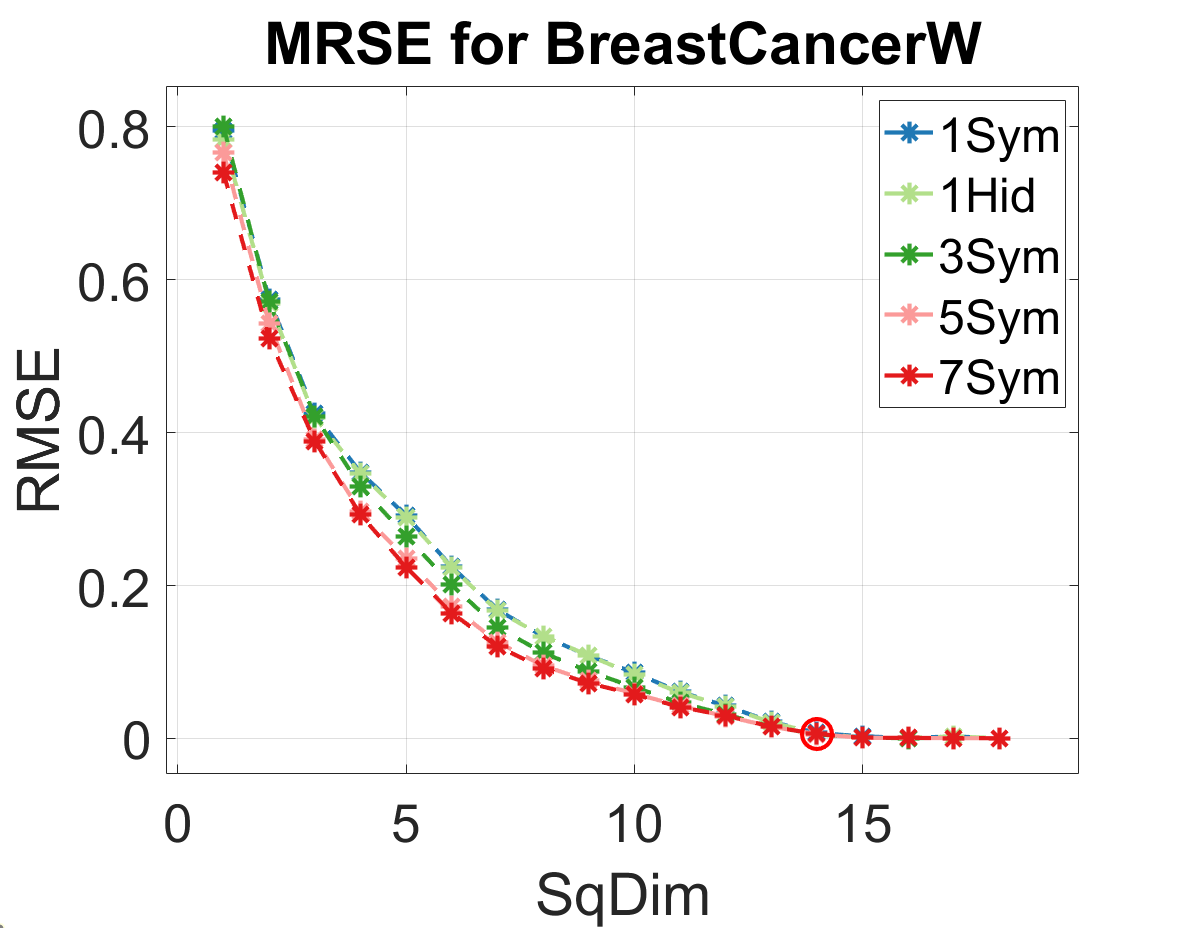}
\caption{\textbf{Left}: RMSEs for BreastCancerW with the original 2-3-4 pattern for the hidden layers. \textbf{Right}: RMSEs for BreastCancerW with 3-5-7 pattern for the hidden layers. Slighly smaller errors were encountered during the early search phase of the larger model on the right but the detected ID and the overall behavior remained the same.}
\label{fig:BreastCancerHidPatComp}
\end{figure*}

\subsection{Generalization of the autoencoder}

In the last experiments, we demonstrate and evaluate the generalization of the additive \nSymFive~autoencoder. Search over squeezing dimensions is performed in a similar manner as that done in Section \ref{subsec:OptExp3}. We apply a small sample of datasets, for which a separate validation data was given in the UCI repository. More precisely, we use Letter (size of training data $N = 16 000$, size of validation data $N_v = 4 000$, i.e., 80\%--20\% portions with respect to the entire data; number of nonconstant featureas $n = 16$), UJIIndoor ($N = 19 937,\ N_v = 1111,$ 95\%--5\% portions; $n = 473$), HumActRecog ($N = 7 351,\ N_v = 2 946,$ 71\%--29\% portions; $n = 561$), and MNIST ($N = 60 000,\ N_v = 10 000,$ 86\%--14\% portions; $n = 666$). Note that because all data are used as is, we have no information or guarantees on how well the data distributions in the training and validation sets actually match each other.

\begin{figure*}[t!]
\centering
\hspace*{\FigsCut}
\includegraphics[width=\figurescale]{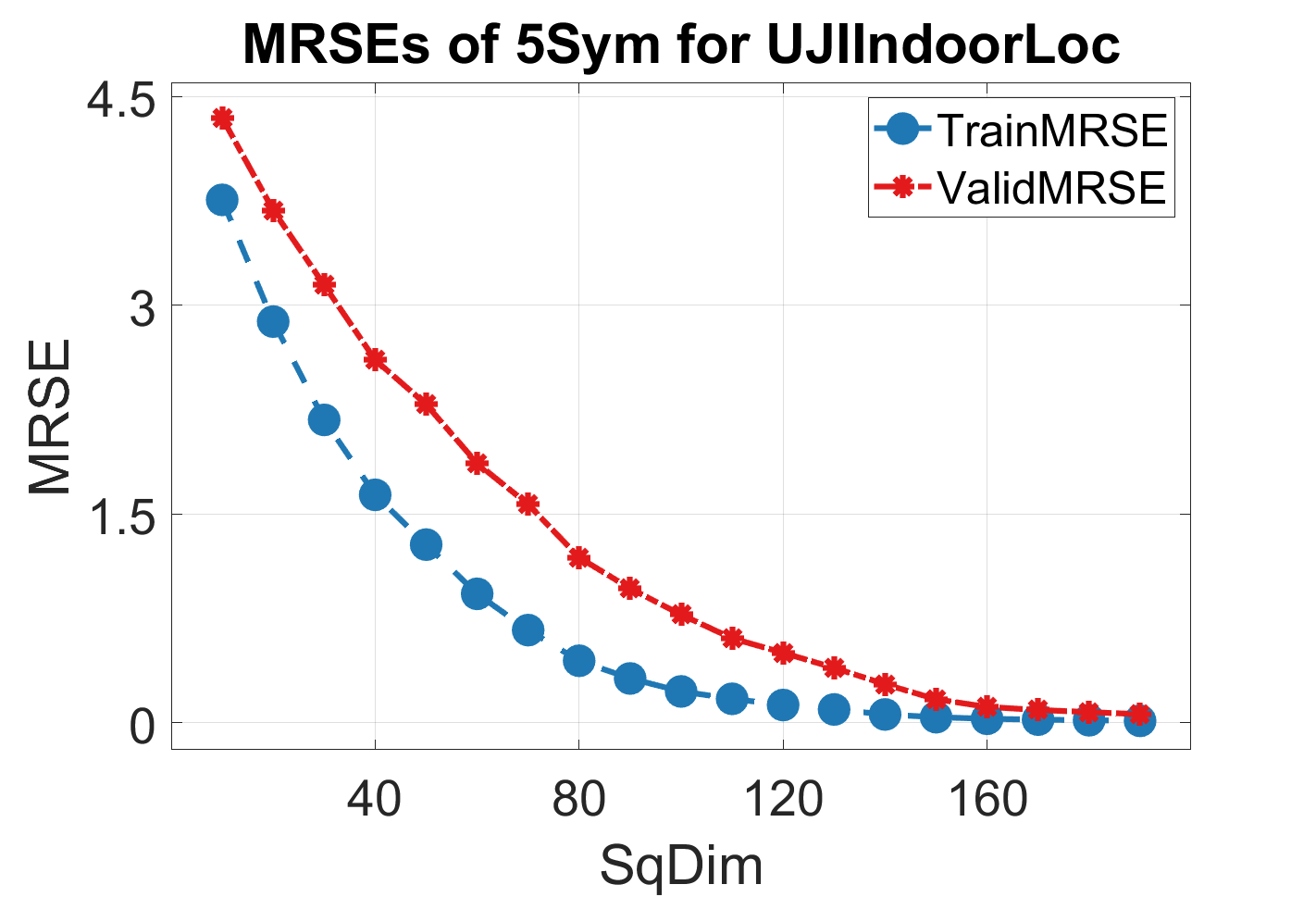}
\includegraphics[width=\figurescale]{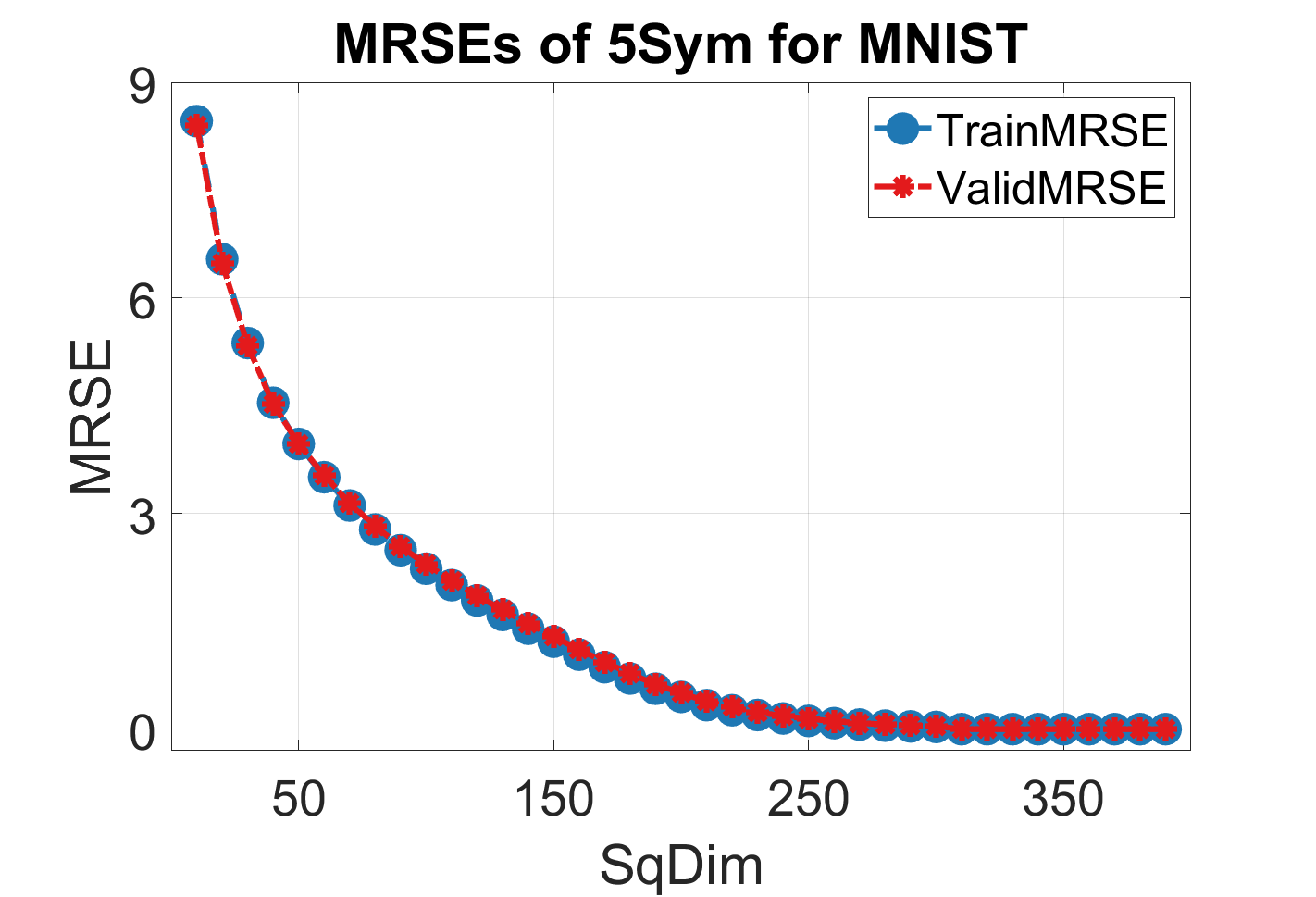}
\caption{Agreements of training and validation set MRSE values for UJIIndoor (left) and MNIST (right). Large deviation between training-validation errors on the left but perfect match on the right. In ID, similar autoencoding error level is reached with both datasets.}
\label{fig:UJIIndMNISTTrainValid}
\end{figure*}

As anticipated, both training-validation portions and the data dimension affected the generalization results. For Letter, with 80\%-20\% division between training-validation sizes and small number of features, we witnessed a perfect match between the training and validation MRSE values. The same held true for MNIST, which is illustrated in Fig. \ref{fig:UJIIndMNISTTrainValid} (right). The largest discrepancy between the training and validation errors, depicted in Figure \ref{fig:UJIIndMNISTTrainValid} (left), was obtained for UJIIndoor, which has the most deviating 95\%-5\% portions with almost 500 features. This dataset also had one of the largest efficiencies (i.e., reduction potential) in Table \ref{tab:RelErrsLargeDatasets}. HumActRec was somewhere in the middle in its behavior, with clearly visible deviation. Because of the data portions ($\sim$70\%-30\%), the difference raises doubts regarding the quality of the validation set. Note that these considerations provide examples of the possibilities of autoencoders to assess the quality of data.

The visual inspection was augmented by computing the correlation coefficient between the MRSE values in the training and validation sets. The following values confirmed the conclusions of the visual inspection: Letter 1.0000, UJIIndoor 0.9766, HumActRecog 0.9939, and MNIST 0.9999. Finally, an important observation from Fig. \ref{fig:UJIIndMNISTTrainValid} is that when the squeezing dimension is increased up to the intrinsic dimension, then the validation error tends to the same error level than the training error. Therefore, the additive autoencoder determined using the training data was always able to explain the variability of the validation data with a compatible accuracy.

\section{Conclusions}\label{sec:Conclu}

This study illustrated a case where all main concerns with feedforward mappings summarized in \cite[p.~363]{hornik1989multilayer} were solved: learning was successful, the size and the number of hidden layers were identified, and the deterministic relationship within a dataset was found. Similar to \cite{hinton2006reducing}, stacking was found to be an essential building block for estimating the weights of deep autoencoders. However, the pretraining face was conceptually (the structure and optimization of the weights) one-to-one with the corresponding part of the final, fine-tuned autoencoder. This was different from  deep residual networks \citep{he2016deep}, where layer skips over multiple layers with batch normalization were applied. The other main ingredients of the proposed autoencoder were an automatically scalable cost function with compact layerwise weight calculus 
and
a simple heuristic for determining the intrinsic data dimension.
Intrinsic dimensions for all tested datasets with a low autoencoding error were revealed. A similar autoencoding error, and the corresponding intrinsic dimension, was obtained independently on the depth of the network. 

One clear advantage of the proposed methodology is the lack of meta-level parameters (e.g., number and form of layers, selection of activation function, detection of the learning rate) that are usually tuned or grid-searched when DNNs are trained. The only parameter that may need adjustment based on visual assessment is $\tau$\textemdash that is, the threshold for identifying the hidden dimension. Moreover, because of the observed smoothly decreasing behavior of the autoencoding error, the intrinsic dimension could be searched for more efficiently than just incrementally: One could attempt to utilize one-dimensional optimization techniques like a golden-section search and/or polynomial and spline interpolation to more quickly identify the beginning of the error plateau.

These results challenge the common beliefs and currently popular traditions with deep learning techniques. 
The experiments summarized here and given in the SI suggest that many existing deep learning results could be improved by using a clear separation of linear and nonlinear data-driven models. Also use of more accurate optimization techniques to determine the weights of such models may be advantageous. 


We can use the additive transformation to the intrinsic dimension as a pretrained part for transfer learning with any prediction or classification model \citep{ghods2021survey}. It would be interesting to test in the future whether one should use this as is or would a transformation into a smaller squeezing dimension than the intrinsic one generalize better in prediction and classification tasks? Another detectable dimensions of the squeezing layer worth investigating, as illustrated in the relative MRSE plots (see also the SI) and in Tables \ref{tab:RelErrsSmallDatasets} and \ref{tab:RelErrsLargeDatasets}, could be the one with the largest nonlinear gain\textemdash that is, with the maximum difference between the PCA error and the autoencoder error or between the shallow and deep results. Moreover, we used global techniques in every part of the autoencoder. The technique might benefit from encoding locally estimated behavior\textemdash for example, using convolutional layers for local-level estimation \citep{lecun1990handwritten}. Similarly, other linear transformation techniques and modifications of PCA might provide better performance  \citep{burges2010dimension,song2018brief,vogelstein2021supervised}, although in the proposed form, we also need the inverse of the linear mapping to be able to estimate the residual error in the original vector space.

\section*{Supplementary information and availability of data and implementations}
Further information concerning this manuscript is given in the public git-repository:\newline 
\url{https://github.com/TommiKark/AdditiveAutoencoder}.\newline 
There, complementary background material and some additional methodological comparisons are given in a separate Supplementary Information document. Especially, full coverage of all the results and illustrations of the computational experiments are documented there. All data or links for downloading data from public repositories used in the experiments are provided. The reference implementation of the proposed methods are also included.

\section*{Acknowledgments}
This work was supported by the Academy of Finland from the project 351579 (MLNovCat).


\end{document}